\documentclass[11pt]{article}

\usepackage{graphicx}
\graphicspath{{figures/}}

\usepackage{enumitem}

\usepackage{booktabs}
\usepackage{tabularx}
\usepackage{multirow}
\usepackage{array}
\usepackage{float}

\usepackage[hidelinks]{hyperref}
\usepackage{url}

\usepackage{longtable}

\usepackage{caption}

\newcolumntype{L}[1]{>{\raggedright\arraybackslash}p{#1}}

\title{A Systematic Evaluation Protocol of Graph-Derived Signals for Tabular Machine Learning}

\author{
Mario Heidrich$^{1,2}$ \and
Jeffrey Heidemann$^{2}$ \and
Rüdiger Buchkremer$^{2}$ \and
Gonzalo Wandosell Fernández de Bobadilla$^{1}$
}

\date{} 

\begin{document}

\maketitle

\begin{center}
{\small
$^{1}$ Universidad Católica San Antonio de Murcia (UCAM), Murcia, Spain\\
$^{2}$ Institute of IT Management and Digitization Research (IFID), FOM University of Applied Sciences in Economics and Management, Düsseldorf, Germany\\
Correspondence: \texttt{mheidrich@alu.ucam.edu}
}
\end{center}

\begin{abstract}
While graph-derived signals are widely used in tabular learning, existing studies typically rely on limited experimental setups and average performance comparisons, leaving the statistical reliability and robustness of observed gains largely unexplored. Consequently, it remains unclear which signals provide consistent and robust improvements. This paper presents a taxonomy-driven empirical analysis of graph-derived signals for tabular machine learning. We propose a unified and reproducible evaluation protocol to systematically assess which categories of graph-derived signals yield statistically significant and robust performance improvements. The protocol provides an extensible setup for the controlled integration of diverse graph-derived signals into tabular learning pipelines. To ensure a fair and rigorous comparison, it incorporates automated hyperparameter optimization, multi-seed statistical evaluation, formal significance testing, and robustness analysis under graph perturbations. We demonstrate the protocol through an extensive case study on a large-scale, imbalanced cryptocurrency fraud detection dataset. The analysis identifies signal categories providing consistently reliable performance gains and offers interpretable insights into which graph-derived signals indicate fraud-discriminative structural patterns. Furthermore, robustness analyses reveal pronounced differences in how various signals handle missing or corrupted relational data. These findings demonstrate practical utility for fraud detection and illustrate how the proposed taxonomy-driven evaluation protocol can be applied in other application domains.
\end{abstract}


\section{Introduction}
\label{sec:introduction}


Graph-structured data provide rich contextual information that is typically not captured by classical tabular machine learning pipelines. Such data arise naturally in many real-world applications, including fraud detection, social networks, biological networks, and knowledge graphs. While classical tabular machine learning models are typically developed under an independent and identically distributed (i.i.d.) assumption and are widely used in practice, violations of this assumption can distort analysis and modeling when relational dependencies are ignored~\cite{Bishop.2006}. Consequently, modern tabular pipelines rely on feature engineering to compensate for missing relational structure~\cite{Ryan.2021}. For graph-structured data, graph-derived signals are quantitative representations extracted from graph structure and provide a principled way to model these dependencies and enrich machine learning pipelines with relational information~\cite{Hamilton.2020}.

This shift toward graph-derived signals places them squarely within modern tabular learning pipelines, where predictive performance is increasingly driven by representation quality rather than model complexity. Recent benchmarks in tabular machine learning underscore that predictive performance is often governed more by a model's inductive biases and the quality of its input features than by architectural complexity alone~\cite{Grinsztajn.2022}. A rigorous data-centric evaluation by Bartelt et al.~\cite{Bartelt.10.12.202415.12.2024} reinforces this perspective, demonstrating that after expert, dataset-specific feature engineering, performance gaps between advanced models shrink considerably, shifting the critical focus from model selection to representation quality.

Despite this critical insight, the evaluation and application of graph-derived signals often lack a similarly principled, data-centric foundation. Their empirical benefit is frequently assessed through limited, ad hoc experimental setups that report average performance improvements over a narrow set of configurations. Such narrowly defined benchmarks risk drawing biased conclusions, as model rankings and reported gains can be highly sensitive to specific preprocessing choices and evaluation protocols~\cite{Bergstra.2013}. This practice leaves key questions unanswered: Are reported gains statistically reliable and reproducible across random seeds? Are they robust to variations in graph structure or sampling? Furthermore, given the vast and heterogeneous space of possible graph-derived signals, it is often unclear a priori which graph-derived signals are most informative for a specific downstream application task. Consequently, a shift toward evaluation protocols that explicitly assess robustness and statistical reliability is crucial for a meaningful comparison of these learning approaches~\cite{Bartelt.10.12.202415.12.2024}. This lack of systematic evaluation complicates principled model design and poses challenges for drawing reliable conclusions in the field.

To enable reliable progress, we argue that systematic analyses must (i) compare different types of graph-derived signals under a unified evaluation protocol, (ii) explicitly account for randomness in data splits and model training, and (iii) assess whether observed improvements are statistically significant and robust to structural changes in the underlying graph rather than incidental. In the absence of such analyses, practitioners face uncertainty when selecting graph-derived signals from a large and heterogeneous design space for downstream tabular learning tasks.

To navigate this vast design space, a taxonomy-driven perspective is essential. It enables graph-derived signals to be grouped according to the type of graph information they encode. Such an organization supports principled comparison, improves interpretability of empirical results, and facilitates task-specific signal selection across a large design space.

To address these challenges, we conduct a systematic, taxonomy-driven analysis of graph-derived signals in tabular machine learning. Specifically, we evaluate representative graph signal types across multiple information categories using a reproducible experimental protocol that incorporates multi-seed evaluation, robustness analysis under graph perturbations, and formal hypothesis testing. Rather than ranking individual methods, the objective of this study is to assess the statistical relevance and robustness of different classes of graph-derived signals with respect to specific downstream application tasks under controlled conditions. The core contribution of this work is a statistically grounded evaluation protocol that enables reliable, interpretable, and robustness-aware comparison of graph-derived signals for tabular learning.

We demonstrate the proposed evaluation protocol through a practical case study in cryptocurrency transaction fraud detection. While the quantitative findings are necessarily specific to this domain, this application serves as a representative and challenging benchmark to illustrate the general applicability of the proposed evaluation protocol. In this fraud detection setting, relational dependencies between entities encode information that is not accessible to transaction-level features alone, as illicit activity often propagates through complex chains of interactions. By instantiating the methodology on this high-stakes task, we demonstrate how statistically grounded insights can be obtained within a concrete graph-augmented tabular learning setting.

Our evaluation is conducted on the Elliptic Bitcoin Dataset~\cite{Weber.2019}, a commonly used large-scale, temporal, and highly imbalanced benchmark that reflects key challenges of real-world fraud detection. Within this setting, we analyze different categories of graph-derived signals, including proximity-based signals that capture local neighborhood information and structural signals that encode broader role-based patterns, to assess when and how relational information contributes measurable value beyond tabular baselines that do not incorporate graph-derived signals.

Beyond serving as a methodological demonstration, this case study enables a statistically grounded assessment of the practical utility of graph-derived signals in a security-critical application. By combining multi-seed evaluation with formal significance testing, we identify which categories of graph-derived signals yield consistent and statistically reliable performance gains in this setting. Moreover, the taxonomy-driven analysis facilitates interpretation by linking different signal categories to characteristic fraud-related patterns, such as the propagation of illicit activity through local neighborhoods. Finally, robustness experiments under controlled graph perturbations reveal substantial differences in how signal categories degrade under missing or corrupted relational data, highlighting robustness as a key criterion for practical signal selection.

Through this case study, we address the following research questions to investigate the utility, robustness, and interpretability of graph-derived signals within a representative tabular learning setting:

RQ1 (Utility): Do graph-derived signals provide statistically significant performance improvements over tabular baselines that do not incorporate explicit graph-derived information, and does this effect vary across different categories of graph information?

RQ2 (Robustness): How robust are the performance gains from different graph signal categories to controlled degradations (e.g., random edge removal) in the underlying graph structure?

RQ3 (Taxonomy Guidance): To what extent does a taxonomy-driven organization of graph signals facilitate interpretable, category-level insights that can inform signal selection in graph-augmented tabular learning?

\section{Methodology: Evaluation Protocol for Graph Signal Analysis}
\label{sec:methodology}

We propose a reproducible and configurable evaluation protocol designed to assess the statistical impact of graph-derived signals when integrated into classical tabular machine learning pipelines, with a design that supports reuse across transactional and non-transactional graph learning tasks. The proposed evaluation protocol is task-agnostic and enables systematic comparisons between baseline models and models augmented with graph-derived signals.

The evaluation protocol supports the flexible selection of subsets of graph-derived signals drawn from a predefined taxonomy of graph information types. In our experiments, we consider a diverse set of graph-derived signals, including features and embeddings spanning centrality, cohesion, community, proximity, spectral, structural role, and neighborhood-based information (see Section~\ref{sec:graph-signal-taxonomy}).

To ensure statistically reliable conclusions, the evaluation protocol explicitly accounts for randomness in both data splitting and model training. Multiple random seeds are applied consistently across baseline and graph-augmented configurations, enabling run-level comparisons. The evaluation protocol further supports experiments on multiple graph variants derived from the same underlying entities, such as perturbed graphs obtained through edge removal, allowing robustness analyses under structural changes.

In addition to standard multi-seed evaluation, the evaluation protocol explicitly supports robustness analysis under controlled structural perturbations of the input graph. In this work, robustness is assessed by systematically removing a predefined fraction of edges prior to graph signal generation, allowing the sensitivity of different graph signal categories to degraded graph structure to be analyzed under otherwise identical experimental conditions.

For fair model comparison, all classifiers are independently optimized using automated hyperparameter optimization based on Bayesian optimization using the Tree-structured Parzen Estimator~\cite{Bergstra.2011}.

Optimization is performed using cross-entropy loss, which is well suited for highly imbalanced classification settings. Depending on the classifier, additional mechanisms such as class weighting, resampling strategies, feature normalization, embedding-specific dimensionality reduction, architectural choices (where applicable), and model-specific regularization are considered during optimization. Importantly, all preprocessing steps are optimized in a model-aware manner and fitted exclusively on the training data to avoid information leakage.

All optimized models are evaluated on identical train–validation–test splits, ensuring comparability across feature configurations. Detailed results are stored in a structured format that enables automated downstream analysis. In particular, the evaluation protocol supports:
\begin{enumerate}[label=(\roman*)]
	\item aggregation of performance metrics across random seeds using trimmed aggregation to reduce the influence of outlier runs, 
	\item stability analysis via standard deviation estimates,
	\item statistical hypothesis testing to assess whether observed performance differences between baseline and graph-augmented models are statistically significant, and 
	\item the computation and storage of model-specific feature importance scores for post-hoc analysis.
\end{enumerate}
Graph-derived signals are concatenated with task-specific base features, resulting in an augmented tabular representation that can be processed by standard classifiers.

\textbf{Trimmed Performance Aggregation}

To reduce the influence of outlier runs and obtain robust performance estimates, we aggregate $F_1$-scores across random seeds separately for each classifier and each graph signal using a trimmed mean approach.

Specifically, for each fixed classifier–signal configuration $(c,g)$, we collect $S=10$ $F_1$-scores per random seed for both the baseline (transaction-only) model and the corresponding graph-augmented model, where each seed corresponds to one complete experimental run with identical data split and training conditions. In total, the evaluation considers $G=24$ distinct graph signals. All analyses exclude Naive Bayes, as motivated in Section~\ref{subsec:results-overview}.

Let $F^{(s)}(c,g),\; s \in \{1,\dots,S\}$ denote the $F_1$-score obtained for classifier $c$ with graph signal $g$ under random seed $s$.

Analogously, let $F_1^{(s)}(c,\mathrm{base})$ denote the $F_1$-score of the corresponding classifier-specific baseline model under the same seed.

To obtain robust performance estimates, we apply a trimmed mean over the $S$ seed-level scores by removing the minimum and maximum values and averaging the remaining $S-2$ runs. Formally, the trimmed mean $F_1$-score for a given classifier–signal configuration is defined as

\[
\bar{F}_1^{\mathrm{trim}}(c,g)
=
\frac{1}{S-2}
\sum_{s \in S_{\mathrm{mid}}}
F_1^{(s)}(c,g)
\]

where $S_{\mathrm{mid}}$ denotes the index set of the remaining runs after excluding the lowest and highest $F_1$-scores across seeds.

The trimmed baseline performance $\bar{F}_1^{\mathrm{trim}}(c,\mathrm{base})$ is computed analogously.

The performance gain induced by graph signal $g$ for classifier $c$ is then computed as the difference between the trimmed mean performances:
\[
\Delta F_1(c,g)
=
\bar{F}_1^{\mathrm{trim}}(c,g)
-
\bar{F}_1^{\mathrm{trim}}(c,\mathrm{base})
\]

Positive values of $\Delta F_1(c,g)$ indicate an improvement over the transaction-only baseline, while negative values indicate degraded performance. This aggregation strategy yields a robust central tendency estimate that mitigates the influence of extreme seed-dependent variations while preserving paired comparability between baseline and graph-augmented models.

All reported $F_1$-score averages and performance differences in Section~\ref{sec:experimental-results} and the Appendices are computed using this trimmed aggregation across random seeds.

\textbf{Statistical Significance Testing}

To assess whether observed performance differences are statistically significant, we apply McNemar’s test to paired predictions obtained from baseline (transaction-only) and graph-augmented models evaluated on identical test splits. Statistical significance is assessed at the prediction level, while performance differences are reported in terms of aggregated $F_1$-scores as described in the previous section.

For a given classifier c, graph signal g, and random seed s, the test compares discordant prediction outcomes between the baseline and the corresponding graph-augmented model. Let b denote the number of instances where the baseline prediction is correct and the graph-augmented prediction is incorrect, and let c denote the number of instances where the baseline prediction is incorrect and the graph-augmented prediction is correct. The McNemar test statistic with continuity correction is given by

\[
\chi^2
=
\frac{(|b - c| - 1)^2}{b + c} .
\]

Differences with $p \leq 0.05$ are considered statistically significant. The test is applied separately for each of the $S=10$ random seeds, where each seed corresponds to one complete experimental run with a fixed data split and model initialization. This ensures that statistical comparisons are fully paired at the prediction level and isolate the effect of incorporating graph-derived signals.

Rather than aggregating p-values across seeds, we summarize statistical evidence across runs by counting, for each classifier–signal combination, how often a graph-augmented model yields a statistically significant improvement versus a statistically significant deterioration relative to the baseline across the $10$ seeds. Specifically, for each $(c,g)$ pair, we record the number of seeds in which the McNemar test indicates a significant improvement $(c > b)$ and the number of seeds indicating a significant deterioration $(b > c)$. These counts form the basis for the aggregated significance analyses and visualizations reported in Section~\ref{sec:experimental-results}.

This aggregation strategy preserves seed-level statistical validity while providing an interpretable summary of the consistency and directionality of statistically significant effects across independent experimental runs.

\textbf{Signal Concatenation and Combined Evaluation}

The evaluation protocol supports not only the evaluation of individual graph signals but also arbitrary combinations of signals through concatenation. For any subset of graph signals $G \subseteq \{g_1,\dots,g_{24}\}$ drawn from the predefined taxonomy, we define the concatenated feature vector for node v as:

\[
x_v^{(g)}
=
\bigl[
x_v^{\mathrm{base}}
\;\|\;
x_v^{(g_1)}
\;\|\;
\cdots
\;\|\;
x_v^{(g_{|G|})}
\bigr]
\]

where $x_v^{\mathrm{base}}$ denotes the original transaction features, $x_v^{(g)}$ the feature vector of graph signal $g$, and $\|$ the concatenation operator. This design enables the evaluation of:
\begin{itemize}
	\item individual graph signals ($|G| = 1$),
	\item category-level combinations (e.g., all centrality-based signals), and
	\item larger multi-signal combinations spanning multiple categories.
\end{itemize}
All evaluation metrics and statistical tests described above are applied identically to both individual and concatenated signal configurations, ensuring consistent comparison across granularity levels. In Section~\ref{sec:experimental-results}, we report results for both individual graph signals and category-level concatenations; for example, ‘Centrality’ denotes the concatenation of all five centrality-based indicators listed in Table~\ref{tab:graph-signals-taxonomy}.

\section{Related Work}
\label{sec:related-work}

\subsection{Classical Graph Indicators}
\label{subsec:classical-graph-indicators}

Classical graph indicators have long been used as hand-crafted features to characterize node importance and local connectivity patterns in complex networks. Common examples include degree-based measures, centrality metrics such as PageRank and betweenness centrality, as well as clustering coefficients capturing local transitivity. These features provide interpretable summaries of node-level structure and have been widely applied in graph-based classification and fraud detection settings (see, e.g., Network Science~\cite{Barabasi.2016}).

Community detection methods extend this perspective by capturing mesoscopic graph structure through partitioning nodes into densely connected groups. Algorithms such as Infomap~\cite{Rosvall.2008} and the Leiden algorithm~\cite{Traag.2019} have been shown to produce stable and high-quality community assignments and are frequently used to derive node-level community features in large transaction networks.

\subsection{Proximity-Based Node Embedding Methods}
\label{subsec:proximity-embeddings}
Unsupervised node embedding methods aim to learn continuous vector representations that preserve graph proximity and local neighborhood relationships. Random-walk-based approaches such as DeepWalk~\cite{Perozzi.08242014} and node2vec~\cite{Grover.2016} generate node sequences via truncated random walks and learn embeddings by modeling node co-occurrence. These methods represent a widely adopted class of unsupervised embedding techniques for learning node representations from the graph.

\subsection{Structural and Role-Oriented Embedding Methods}
\label{subsec:structural-role-embeddings}
Beyond proximity preservation, a class of approaches explicitly focuses on capturing structural similarity or node roles, independent of local neighborhoods. These methods aim to group nodes with similar structural functions rather than relying on direct neighborhood overlap.

These approaches primarily differ in how structural similarity is operationalized and encoded through their algorithmic design choices. Representative examples include GraphWave~\cite{Donnat.07192018}, which leverages spectral graph wavelets to encode structural signatures across multiple diffusion scales, as well as role-based embedding methods such as role2vec~\cite{Ahmed.2022} and ffstruc2vec~\cite{Heidrich.2025}, which model structural identity by capturing similarities in node roles.

\subsection{Spectral Graph Embeddings}
\label{subsec:spectral-embeddings}
Spectral embedding techniques derive node representations from eigenvectors of the graph Laplacian, providing a principled global view of graph structure. By projecting nodes into a low-dimensional space spanned by non-trivial eigenvectors, spectral methods capture connectivity patterns beyond immediate neighborhoods. While computationally demanding for large graphs, spectral embeddings remain a foundational approach in graph representation learning (e.g., Spectral Graph Theory~\cite{Chung.2009}).

\subsection{Graph Neural Networks and Graph Contrastive Learning}
\label{subsec:gnn-gcl}
Neighborhood-based aggregation is commonly realized through Graph Neural Networks (GNNs), which iteratively propagate and aggregate information from local neighborhoods to learn node representations. Canonical instances include Graph Convolutional Networks (GCN), which employ normalized linear aggregation of neighboring features~\cite{Kipf.2017}, and Graph Attention Networks (GAT), which extend this paradigm by learning adaptive, attention-based weighting schemes over neighbors~\cite{Velickovic.2018}. These architectures have demonstrated strong performance across a wide range of node classification tasks.

More recently, Graph Contrastive Learning (GCL) has emerged as a self-supervised representation learning paradigm that aims to learn robust node embeddings by contrasting different augmented views of graph data. Representative approaches, such as BGRL-style contrastive learning~\cite{Thakoor.2021}, optimize contrastive objectives without relying on labeled supervision and have shown promising results in downstream tasks.

\section{Taxonomy of Graph Signals for Tabular Machine Learning}
\label{sec:graph-signal-taxonomy}
Graph-structured data contain rich relational information that can significantly enhance predictive models when integrated into classical tabular machine learning pipelines. As discussed in the previous section, such relational dependencies motivate the use of graph-derived signals alongside traditional tabular features~\cite{Hamilton.2020}. However, the space of potential graph signals is vast and heterogeneous, ranging from simple local statistics to complex learned embeddings. In this work, we use the term graph signal to denote any quantitative representation derived from the graph structure that can be associated with nodes and integrated into downstream tabular learning models, including hand-crafted indicators, unsupervised embeddings, and learned neighborhood-based representations. Without a structured conceptual basis for selecting and evaluating these signals, practitioners face a signal selection problem, as it is often unclear which types of graph-derived signals are most relevant for a given downstream task.

Over the past years, a wide range of graph-based methods has been proposed, capturing different aspects of network structure. Rather than treating these methods as isolated techniques, it is useful to organize them according to the type of information they extract from the graph. This section provides a concise, non-exhaustive taxonomy of graph-related information that can be incorporated into tabular machine learning models, serving as a conceptual foundation for the feature selection used in this study.

The taxonomy presented in this section builds on well-established concepts and recent work in graph representation learning and network analysis \cite{Barabasi.2016,Rosvall.2008,Traag.2019,Perozzi.08242014,Grover.2016,Donnat.07192018,Ahmed.2022,Heidrich.2025,Chung.2009,Kipf.2017,Velickovic.2018,Thakoor.2021}. It consolidates commonly used categories of graph-derived signals and serves as a structured conceptual basis for the comparative evaluation conducted in this study.

We distinguish six broad categories of graph-based information, reflecting different structural perspectives on the same underlying graph: centrality, cohesion, community structure, proximity, spectral properties, and structural role information. These categories are conceptually distinct but complementary, and together cover a wide spectrum of graph signals commonly used in practice. Neighborhood-based features, as exploited by message-passing architectures such as Graph Neural Networks, can be viewed as an aggregation mechanism that combines multiple categories rather than as a separate information type.

\subsection{Centrality-Based Indicators}
\label{subsec:centrality-indicators}
Centrality-based indicators quantify the importance or influence of nodes within the network. Typical examples include degree-based measures, PageRank, betweenness, closeness, or eigenvector centrality. Such indicators capture how prominently a node is positioned within the overall connectivity structure and often provide strong baseline signals for downstream classification tasks.

\subsection{Cohesion-Related Indicators}
\label{subsec:cohesion-indicators}
Cohesion-related indicators describe the local interconnectedness of a node’s neighborhood. Measures such as clustering coefficients, core numbers, or triangle counts capture the extent to which a node participates in tightly connected substructures. These signals are particularly relevant in domains where meaningful differences in local connectivity patterns reflect distinct behavioral or structural patterns in the graph.

\subsection{Community-Based Indicators}
\label{subsec:community-indicators}
Community-related indicators focus on meso-scale structures in the graph. Community detection methods assign nodes to groups of densely connected subgraphs, providing coarse-grained structural context. Community membership indicators can serve as high-level features that summarize a node’s structural environment beyond immediate neighborhoods.

\subsection{Proximity-Based Embeddings}
\label{subsec:proximity-embeddings-taxonomy}
Proximity-based embeddings encode relative closeness between nodes in a latent space, reflecting similarity induced by graph connectivity. Techniques such as DeepWalk, node2vec, or related approaches can be used to derive such proximity-based representations such that nodes with similar connectivity patterns or overlapping neighborhoods are placed close together in the representation space. These embeddings are effective at capturing distance-based similarity and local context.

\subsection{Spectral Graph Embeddings}
\label{subsec:spectral-embeddings-taxonomy}
Spectral information derives from the eigenstructure of graph-related matrices, such as the adjacency matrix or the graph Laplacian. Spectral features encode global connectivity patterns and can capture structural regularities that are not easily observable through local measures alone.

\subsection{Structural Role Embeddings}
\label{subsec:structural-role-embeddings-taxonomy}
Structural role information aims to characterize nodes according to their position within the global topology of the graph, independent of direct proximity. Structural embedding methods focus on identifying nodes that occupy similar roles—such as hubs, bridges, or peripheral nodes—even if they are located far apart, as exemplified by approaches such as struc2vec, GraphWave, role2vec, and ffstruc2vec.

\subsection{Graph Neural Networks and Graph Contrastive Learning}
\label{subsec:gnn-gcl-taxonomy}
Graph Neural Networks (GNNs) and recent Graph Contrastive Learning (GCL) approaches constitute a powerful class of learning mechanisms that operate on graph-structured data by aggregating and transforming information from node neighborhoods. In the proposed taxonomy, these methods are viewed as integrative representation learning paradigms rather than as individual graph signal types.

While these architectures are inherently capable of end-to-end classification, we employ them within our evaluation protocol specifically as representation learners to derive high-dimensional node embeddings, which serve as signals for the subsequent tabular analysis.

\subsection{Selection of Graph Signals and Role of the Taxonomy}
\label{subsec:signal-selection-taxonomy}
In this work, we select a set of 24 representative graph-derived signals, with multiple representatives drawn from each category of the proposed taxonomy (see Table~\ref{tab:graph-signals-taxonomy}). The selection is guided by representativeness with respect to the underlying information category. The goal is to assess how different types of graph signals contribute to predictive performance when integrated into tabular machine learning models.

This taxonomy provides the conceptual basis for interpreting the experimental results presented in the following sections. In particular, it allows performance differences between feature groups to be analyzed at the level of graph signal categories, while also enabling more fine-grained comparisons between individual graph signals within the same category. This dual perspective offers insights into which types of graph signals are most beneficial for the application scenario under consideration.

In our experimental setup, GNN- and GCL-based representations are treated consistently with other graph signals by extracting fixed node embeddings from the final encoder layer. While GCL is trained using an unsupervised contrastive objective, GCN and GAT encoders are optimized in a supervised fashion on the node classification task. To ensure a fair evaluation and prevent data leakage, all encoder training is strictly confined to the respective training splits. For GCL, we use an unsupervised contrastive learning setup, and the resulting embeddings are concatenated with transaction-level features and processed by the same tabular classifiers as all other graph-derived signals.

All GNN-based representations are generated via a two-stage procedure consisting of encoder training (supervised for GCN/GAT, self-supervised for GCL) followed by frozen embedding extraction, thereby maintaining a clear separation between representation learning and downstream classification. This setup is intentional and enables a comparative assessment of supervised neural representations alongside purely structural and self-supervised graph signals, thereby capturing the maximum attainable performance of each signal class under controlled conditions.

\setlength{\tabcolsep}{6pt}
\renewcommand{\arraystretch}{1.15}

\begin{longtable}{L{3.2cm} L{4.0cm} L{\dimexpr\textwidth-3.2cm-4.0cm-4\tabcolsep\relax}}
\caption{Representative graph signals categorized by taxonomic family.}
\label{tab:graph-signals-taxonomy}\\
\toprule
\textbf{Category} & \textbf{Method} & \textbf{Description} \\
\midrule
\endfirsthead

\toprule
\textbf{Category} & \textbf{Method} & \textbf{Description} \\
\midrule
\endhead

\multirow{5}{*}{Centrality}
& Degree Centrality & Number of incident edges \\
& PageRank & Recursive importance scoring \\
& Betweenness Centrality & Fraction of shortest paths through node \\
& Eigenvector Centrality & Importance based on neighbors' importance \\
& Closeness Centrality & Inverse average shortest path length \\
\midrule

\multirow{4}{*}{Cohesion}
& Clustering Coefficient & Density of direct neighbors \\
& Core Number & Max $k$ such that node belongs to $k$-core \\
& Triangle Count & Number of triangles involving node \\
& Average Clustering & Mean clustering coefficient of neighbors \\
\midrule

\multirow{3}{*}{Community}
& Louvain & Modularity-maximizing community detection \\
& Leiden & Improved Louvain with refinement \\
& Infomap & Information-theoretic community detection \\
\midrule

\multirow{4}{*}{Proximity}
& DeepWalk & Random walks + Skip-gram \\
& node2vec (BFS) & Biased walks emphasizing breadth \\
& node2vec (DFS) & Biased walks emphasizing depth \\
& node2vec (balanced) & Balanced BFS/DFS exploration \\
\midrule

Spectral
& Spectral Embedding & Laplacian eigenmap embeddings \\
\midrule

\multirow{3}{*}{Structural Role}
& ffstruc2vec & Structural identity via flat similarity graph \\
& GraphWave & Spectral graph wavelet embeddings \\
& role2vec & Role-based node embeddings \\
\midrule

\multirow{3}{*}{\parbox[t]{3.2cm}{\raggedright GNN / Neighborhood\\Features}}
& GCN & Graph Convolutional Network \\
& GAT & Graph Attention Network \\
& GCL & Graph Contrastive Learning \\
\bottomrule

\end{longtable}

Graph signals are computed using fixed, standard parameter settings. The study focuses on the comparative evaluation of different types of graph-derived signals.

\section{Case Study: Fraud Detection on Elliptic Bitcoin Transaction Graph}
\label{sec:case-study-elliptic}
The proposed taxonomy and graph signal analysis are evaluated in the context of transaction fraud detection using the Elliptic Bitcoin Dataset, a widely used benchmark for illicit activity detection in cryptocurrency transaction networks~\cite{Weber.2019}.

The Elliptic dataset represents a temporal directed transaction graph, where nodes correspond to Bitcoin transactions and edges represent fund flows between transactions. Each node is associated with a timestamped feature vector and a binary label indicating licit or illicit activity, with a substantial portion of nodes remaining unlabeled. The graph structure introduces strong relational dependencies between samples, violating the independent and identically distributed (i.i.d.) assumption underlying classical tabular learning.

Concretely, the dataset contains 203,769 nodes and 234,355 directed edges, distributed across 49 temporal snapshots. Among the labeled transactions, the class distribution is highly imbalanced, with illicit transactions accounting for roughly 2–3 \% of labeled nodes, reflecting realistic fraud detection conditions. These characteristics make the Elliptic dataset a challenging and representative benchmark for evaluating graph-derived signals under severe class imbalance and structural dependency.

Rather than focusing on a single modeling paradigm, this study investigates how explicitly extracted graph-derived signals contribute to predictive performance when integrated into classical tabular machine learning pipelines.

Instead of aiming to optimize end-to-end performance on the Elliptic benchmark or to propose a task-specific fraud detection model, this case study uses the dataset as a controlled experimental environment to analyze the contribution of explicitly extracted graph-derived signals. The focus is on assessing whether the inclusion of graph-derived signals leads to statistically significant and reproducible performance improvements over a transaction-only baseline, and on understanding how these effects vary across different graph signal categories. By conducting the evaluation on a well-established and widely used benchmark, the results are intended to provide general insights into the utility of different graph signal types, rather than benchmark-specific performance claims.

This experimental framing enables performance differences to be analyzed both across graph signal categories defined by the proposed taxonomy and at a more fine-grained level within individual categories, facilitating a structured interpretation of which types of graph-derived signals—such as centrality, proximity, or structural role information—are most beneficial for fraud detection under realistic conditions.

\section{Experimental Setup}
\label{sec:experimental-setup}
This section describes the concrete experimental setup used to instantiate the proposed evaluation protocol on the Elliptic fraud detection benchmark. We detail the considered classifiers, data splits, evaluation metrics, hyperparameter optimization strategy, robustness settings, and the statistical analysis procedures used to assess the significance of observed performance differences, ensuring full reproducibility and transparency.

\subsection{Prediction Task and Baseline}
\label{subsec:prediction-task-baseline}
The prediction task is binary node classification, where each transaction is classified as licit or illicit based on available transaction-level features. As a baseline, we consider classical tabular machine learning models trained exclusively on the original transaction features provided by the dataset, without incorporating any graph-derived signals.

We exclude the native Elliptic graph features, as they are tailored to this specific benchmark and not intended as general-purpose graph descriptors. Instead, we focus on general-purpose graph-derived signals to enable a controlled and comparable evaluation across methods and settings.

Graph-augmented models extend this baseline by concatenating graph-derived signals with the original transaction features, resulting in an augmented tabular representation processed by the same classifiers. This design enables controlled, paired comparisons between baseline and graph-augmented models.

\subsection{Classifiers}
\label{subsec:classifiers}
To ensure broad coverage of commonly used tabular learning paradigms, we evaluate a diverse set of classifiers with different inductive biases. Specifically, the experimental setup includes linear models, probabilistic classifiers, kernel-based methods, and tree-based ensemble models.

Concretely, we consider Logistic Regression, Naive Bayes, Support Vector Classifiers (SVC) with both linear and RBF kernels, Random Forests, and XGBoost~\cite{Chen.08132016}. This selection covers a wide range of modeling assumptions, including linear decision boundaries, probabilistic independence assumptions, margin-based classification, and non-linear ensemble learning.

All classifiers are trained and evaluated under identical experimental conditions to ensure fair comparison. Each model is optimized independently using automated hyperparameter optimization based on Bayesian optimization with a fixed budget of 50 trials per configuration. No classifier-specific tuning or manual intervention is performed outside this unified optimization framework.

\subsection{Data Splits and Random Seeds}
\label{subsec:data-splits-seeds}
We adopt a transductive evaluation setting following standard protocols in graph machine learning~\cite{Kipf.2017, Yang.2016}, where the full graph structure is assumed to be available during training, while labels are observed only for the respective training nodes. This setting avoids confounding effects caused by strong temporal shifts in label distributions and enables fair, paired comparisons between baseline and graph-augmented models under identical structural conditions.

To account for randomness in both data splitting and model training, we perform all experiments across multiple random seeds. This multi-seed evaluation protocol enables run-level paired comparisons between baseline and graph-augmented models, allowing variability due to randomness to be explicitly quantified rather than implicitly averaged out.

For each seed, the dataset is split into disjoint training, validation, and test sets using a stratified random split with a $60$--$20$--$20$ ratio, and identical split indices are applied across all graph signal configurations.

Hyperparameter optimization is performed once for each classifier and graph signal configuration using a fixed random seed ($seed = 42$) for data splitting and model initialization. The resulting optimal hyperparameters are then fixed and used consistently across all corresponding experiments, which are evaluated over multiple random seeds (seeds 1--10) to assess variability and statistical robustness.

\subsection{Hyperparameter Optimization}
\label{subsec:hyperparameter-optimization}
Hyperparameter optimization is performed separately for each classifier and graph signal configuration using Bayesian optimization with a Tree-structured Parzen Estimator~\cite{Bergstra.2011}, as implemented in the hyperopt framework~\cite{Bergstra.2013}. The optimization objective is cross-entropy loss evaluated on the validation set.

In line with the general framework described in Section~\ref{sec:methodology}, classifier-specific mechanisms such as class weighting, oversampling strategies, feature normalization, and embedding-specific dimensionality reduction are included in the optimization space where applicable.

For each optimization run, a fixed budget of 50 trials is used. The best-performing hyperparameter configuration per classifier and graph signal configuration is selected and subsequently reused across all random seeds to ensure consistent evaluation and to avoid leakage between optimization and testing. To ensure full reproducibility and transparency, the complete hyperparameter search spaces for all classifiers and the fixed parameters for all graph signal generators are provided in Appendix~\ref{app:ClassificationHyperparameters} and Appendix~\ref{app:GraphSignalGenerationParameters}, respectively.

\subsection{Evaluation Metrics}
\label{subsec:evaluation-metrics}
Model performance is evaluated using metrics appropriate for highly imbalanced binary classification. As the primary comparison metric, we report the $F_1$-score, which balances precision and recall and enables direct comparison across classifiers and graph signal configurations.

To assess whether observed performance differences between baseline and graph-augmented models are statistically significant, we apply McNemar’s test~\cite{McNEMAR.1947} to paired predictions on identical test splits, considering results with $p \leq 0.05$ as statistically significant.

Precision and recall are further analyzed to support interpretability and to highlight application-relevant trade-offs in fraud detection.

Performance is aggregated across random seeds using trimmed aggregation to reduce the influence of outlier runs.

\subsection{Robustness Experiments under Graph Perturbations}
\label{subsec:robustness-experiments}
To assess the robustness of graph-derived signals to structural changes, we perform additional experiments on perturbed versions of the transaction graph. Perturbations are introduced by randomly removing a fixed proportion of edges while preserving the original node set.

Specifically, edge removal rates of 25 \% and 50 \% are considered to simulate increasing levels of structural degradation.

Graph-derived signals are recomputed for each perturbed graph variant and evaluated using the same experimental protocol as for the unperturbed graph. This allows the stability of observed performance gains to be analyzed under controlled degradation of graph structure.

\subsection{Implementation Details and Reproducibility}
\label{subsec:implementation-reproducibility}
All experiments are implemented using a unified experimental pipeline that ensures consistent preprocessing, model training, and evaluation across configurations. All random seeds, hyperparameter search spaces, and evaluation scripts are fixed and logged to enable full reproducibility.

All code and configuration files required to reproduce the experiments reported in this paper, including data preprocessing, graph signal generation, model training, evaluation scripts, and robustness experiments under graph perturbations, are publicly available.

The complete experimental setup, including configuration files and result logs, is designed to support automated downstream analysis and statistical evaluation.

The experimental pipeline is implemented in Python, leveraging standard machine learning libraries and graph processing frameworks such as scikit-learn, XGBoost~\cite{Chen.08132016}, PyTorch, the Deep Graph Library~\cite{Wang.2019}, and hyperopt~\cite{Bergstra.2013} for automated hyperparameter optimization.

\section{Experimental Results}
\label{sec:experimental-results}

\subsection{Evaluation Overview and Reading Guide}
\label{subsec:results-overview}

While the empirical evaluation is conducted on a single large-scale fraud detection benchmark, the objective of this section is not to maximize task-specific performance scores, but to systematically investigate the relative utility, statistical reliability, and stability of different graph-derived signal categories under a unified and rigorous evaluation protocol.

This section reports the experimental results obtained by augmenting transaction-based fraud detection models with graph-derived signals. We first provide an aggregated comparison between transaction-only baselines and graph-augmented models across all evaluated classifiers and graph signal configurations, using identical data splits, random seeds, and hyperparameters obtained from a fixed optimization protocol to enable fully paired comparisons. This design enables fully paired, run-level comparisons between baseline and graph-augmented models.

Performance is measured using the $F_1$-score, reported as the trimmed mean and standard deviation over multiple random seeds. All results follow the evaluation protocol described in Section~\ref{sec:experimental-setup}, with hyperparameters fixed based on a configuration-specific optimization procedure using seed $42$, and performance aggregated over ten independent runs (seeds $1–10$) using trimmed $F_1$-score statistics.

All reported random seeds affect only data splitting and model training, including initialization and optimization stochasticity. Graph-derived signals are computed deterministically for a given graph and are held fixed across all runs.

To assess the robustness and statistical relevance of observed performance differences, we additionally apply paired McNemar significance tests at a significance level of $p \leq 0.05$, and summarize results across random seeds by counting statistically significant improvements and deteriorations per classifier–signal configuration.

Results are presented at multiple levels of aggregation. We first analyze overall performance trends across graph signal categories, followed by classifier-specific effects, stability across random seeds, and statistical significance patterns. Graph signals are grouped into the following conceptual categories: Centrality, Cohesion, Community, Proximity, Spectral, and Structural, as well as a separate category of Graph Neural Network (GNN)-based representations. Category-level results are obtained by explicitly concatenating all individual graph signals within a category, as described in Section~\ref{sec:methodology}. In addition, detailed results for all $24$ individual graph signals are reported in Appendix~\ref{app:mean-f1}, with corresponding variability statistics provided in Appendix~\ref{app:std-f1}.

For aggregated performance comparisons across classifiers, we exclude the Naive Bayes classifier from the aggregation and report the average $F_1$-score over the remaining classifiers. Naive Bayes consistently exhibits substantially lower performance in this setting, which is expected given its strong conditional independence assumptions and limited capacity to capture complex relational patterns induced by graph-derived signals. Nevertheless, Naive Bayes results are reported in all tables for completeness and transparency, but excluded from aggregated summaries, as including this classifier would disproportionately distort overall trends without providing additional insight.

Section~\ref{subsec:overall-performance-impact} focuses on aggregated performance trends across graph signal categories. Sections~\ref{subsec:classifier-specific-effects} and \ref{subsec:stability-variance-seeds} then analyze classifier-specific effects and stability across random seeds, respectively. A formal statistical significance analysis is provided in Section~\ref{subsec:statistical-significance}, followed by a robustness study under graph perturbations in Section~\ref{subsec:robustness-graph-perturbations}.

\subsection{Overall Performance Impact of Graph Signals}
\label{subsec:overall-performance-impact}
\begin{figure}[H]
  \centering
  \includegraphics[width=\linewidth]{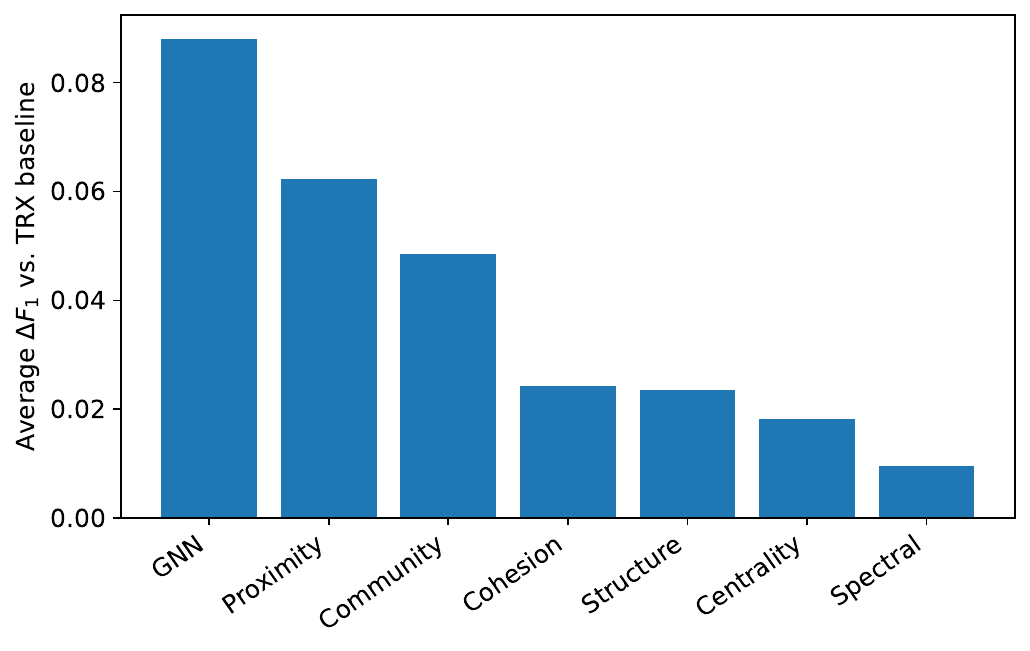}
  \caption{Average $F_1$-score improvements across graph signal categories relative to the transaction-only baseline, aggregated over classifiers and random seeds (trimmed aggregation).}
  \label{fig:cat_delta_vs_trx}
\end{figure}

\begin{figure}[H]
  \centering
  \includegraphics[width=\linewidth]{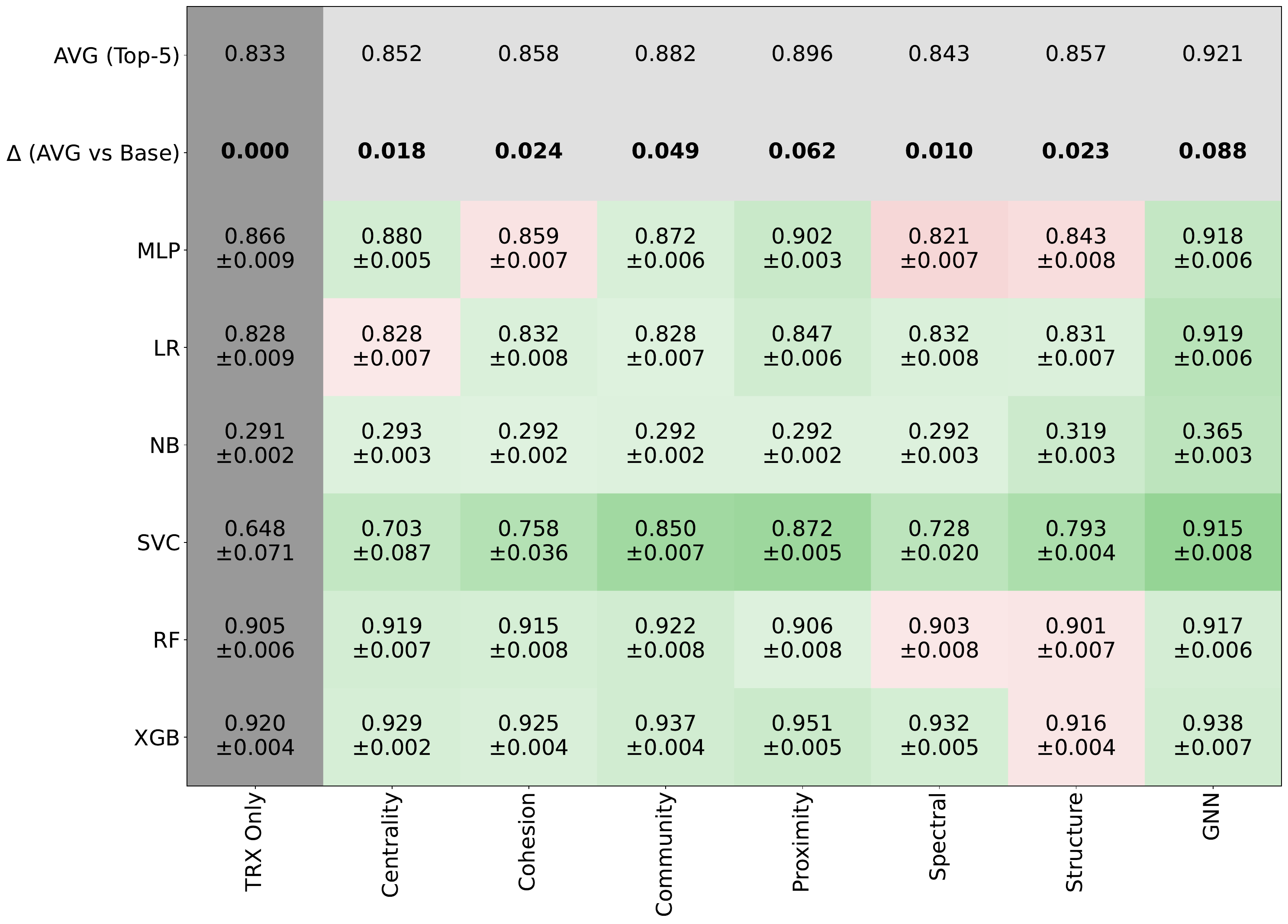}
  \caption{Per-classifier $F_1$-scores across graph signal categories.
  Cell values report mean $F_1$-scores with standard deviations across random seeds.
  Color intensity indicates relative performance differences with respect to the
  transaction-only (TRX) baseline (green = improvement, red = degradation).}
  \label{fig:signal_classifier_matrix}
\end{figure}

Across all evaluated classifiers included in the aggregated analysis, the integration of graph-derived signals leads to consistent performance improvements over the transaction-only baseline. Figure~\ref{fig:cat_delta_vs_trx} summarizes the average $F_1$-score improvements across graph signal categories relative to the transaction-only baseline. Averaged over classifiers and random seeds, most graph signal categories yield positive gains in $F_1$-score, indicating that relational information captured from the transaction network provides complementary predictive power beyond node-local transaction features. The detailed per-classifier performance across all signal categories is shown in Figure~\ref{fig:signal_classifier_matrix} ($F_1$-scores with standard deviations), revealing consistent improvements for most model–signal combinations. For a fine-grained breakdown of results across all $24$ individual graph signals and classifiers, see Appendix~\ref{app:mean-f1} (mean $F_1$-scores) and Appendix~\ref{app:std-f1} (standard deviations across seeds).

Among all evaluated categories, GNN-derived representations achieve the strongest average performance gains. Notably, GNNs are the only class of methods in this study that explicitly incorporate feature information from neighboring nodes during representation learning.

While GNN-derived representations achieve the strongest average performance, an upper-bound analysis in Appendix~\ref{app:PeakPerformanceGainsperGraphSignalCategory} shows that proximity-based graph signals achieve the highest absolute $F_1$-score across all evaluated graph signal categories.

Proximity-based and community-based signals consistently outperform purely structural or spectral representations. This pattern aligns with the intuition that relational closeness and shared transactional context are particularly informative in fraud detection settings, where illicit activity often propagates through localized neighborhoods. Although proximity-based and community-based signals are treated as distinct categories in our taxonomy, both encode aspects of relational closeness at different levels of abstraction. The stronger performance of proximity-based representations suggests that modeling relational proximity in a continuous and fine-grained manner can provide richer information than discrete group assignments in this setting.

Overall, these results demonstrate that graph-based augmentation is generally beneficial in this setting, while also highlighting that performance gains are strongly dependent on the type of relational information being incorporated. In particular, representations that explicitly capture relational proximity and neighborhood context tend to yield the most pronounced improvements, whereas more global or purely structural descriptors provide only limited additional benefit.

While these average performance gains provide a first indication of the benefits of graph-derived signals, their statistical reliability and robustness are examined in the following sections, including a formal significance analysis in Section~\ref{subsec:statistical-significance}.

\subsection{Classifier-Specific Effects}
\label{subsec:classifier-specific-effects}
Building on the aggregated category-level trends shown in Figure~\ref{fig:cat_delta_vs_trx}, which summarize average effects across classifiers, we analyze how graph signal effectiveness varies across individual classifiers, as shown in Figure~\ref{fig:signal_classifier_matrix}, reflecting differences in model capacity and inductive bias.

Before turning to classical classifiers, we briefly discuss the role of GNN-derived representations, which serve as a key source of graph signals in our evaluation. To contextualize the performance of GNN-derived representations, we additionally compare end-to-end GNN models with configurations where fixed node embeddings extracted from the trained GNN encoders are concatenated with transaction-level features and processed by classical tabular classifiers. While end-to-end GNNs achieve strong predictive performance on the Elliptic dataset, the corresponding embedding-based configurations yield higher average $F_1$-scores when combined with optimized tabular models. This observation suggests that, in the evaluated setting, GNNs primarily serve as effective representation learners, whereas downstream tabular classifiers provide more flexible decision boundaries under severe class imbalance. Importantly, this comparison is intended as an interpretative aid rather than a direct architectural comparison, as both approaches are evaluated under different modeling assumptions.

Tree-based ensemble models, particularly Random Forests and XGBoost, exhibit strong and stable performance gains across most graph signal categories. This behavior is consistent with the ability of tree-based ensembles to handle heterogeneous and partially redundant feature sets, which may facilitate the effective integration of complementary graph-derived signals. These models benefit consistently from additional structural features, with improvements observed across centrality, community, proximity, and GNN-based representations. This observation is consistent with recent findings in tabular machine learning, which show that tree-based ensemble models often outperform deep learning approaches when combined with informative feature representations, due to their robustness to heterogeneous feature scales and their ability to exploit complementary feature interactions~\cite{Grinsztajn.2022}.

Support Vector Classifiers show substantial relative improvements over the transaction-only baseline across multiple graph signal categories. However, these improvements are accompanied by increased variability, indicating higher sensitivity to both feature selection and random initialization.

Linear models such as Logistic Regression and shallow neural models (MLP) display moderate but robust improvements, suggesting that even comparatively simple classifiers can exploit graph-derived signals when properly optimized. In contrast, Naive Bayes classifiers show minimal sensitivity to graph augmentation, with performance remaining largely unchanged across signal categories.

Overall, these results highlight that the effectiveness of graph-derived signals depends on the choice of downstream classifier, and that models with higher expressive capacity tend to benefit more consistently from additional relational information.

\subsection{Stability and Variance Across Random Seeds}
\label{subsec:stability-variance-seeds}
To assess the stability of observed improvements, we analyze the variability of $F_1$-scores across random seeds. Across most classifiers and graph signal categories, the standard deviation of performance remains comparable to or lower than the standard deviation observed for the transaction-only baseline. Observed variability is generally low, with standard deviations typically remaining in the range of a few thousandths in $F_1$-score. Figure~\ref{fig:std_f1_signal_matrix} reports the standard deviation of $F_1$-scores across random seeds for individual graph signals and classifiers. Consistent patterns are observed when variability is aggregated at the graph signal category level (see Figure~\ref{fig:cat_delta_vs_trx}), confirming that stability trends persist beyond individual signal realizations. Observed performance improvements generally exceed the corresponding run-to-run variability, indicating that effect sizes are not dominated by random fluctuations.

This observation suggests that performance gains introduced by graph signals are not driven by favorable random initializations or specific data splits. Instead, the consistently low variance across runs indicates that graph augmentation yields reliable and reproducible improvements within the evaluated setting. In several cases, graph augmentation even reduces variability, indicating a stabilizing effect on model behavior.

An exception is observed for Support Vector Classifiers combined with certain centrality-based signals, which exhibit increased variance. Nevertheless, the overall pattern indicates that improvements induced by graph-derived signals are generally robust and reproducible across runs.

\subsection{Statistical Significance Analysis}
\label{subsec:statistical-significance}
To statistically validate the observed performance gains, we conduct paired McNemar tests to assess whether observed improvements are statistically significant across random seeds, following the paired, run-level testing protocol described in Section~\ref{sec:methodology}.

\begin{figure}[H]
  \centering
  \includegraphics[width=\linewidth]{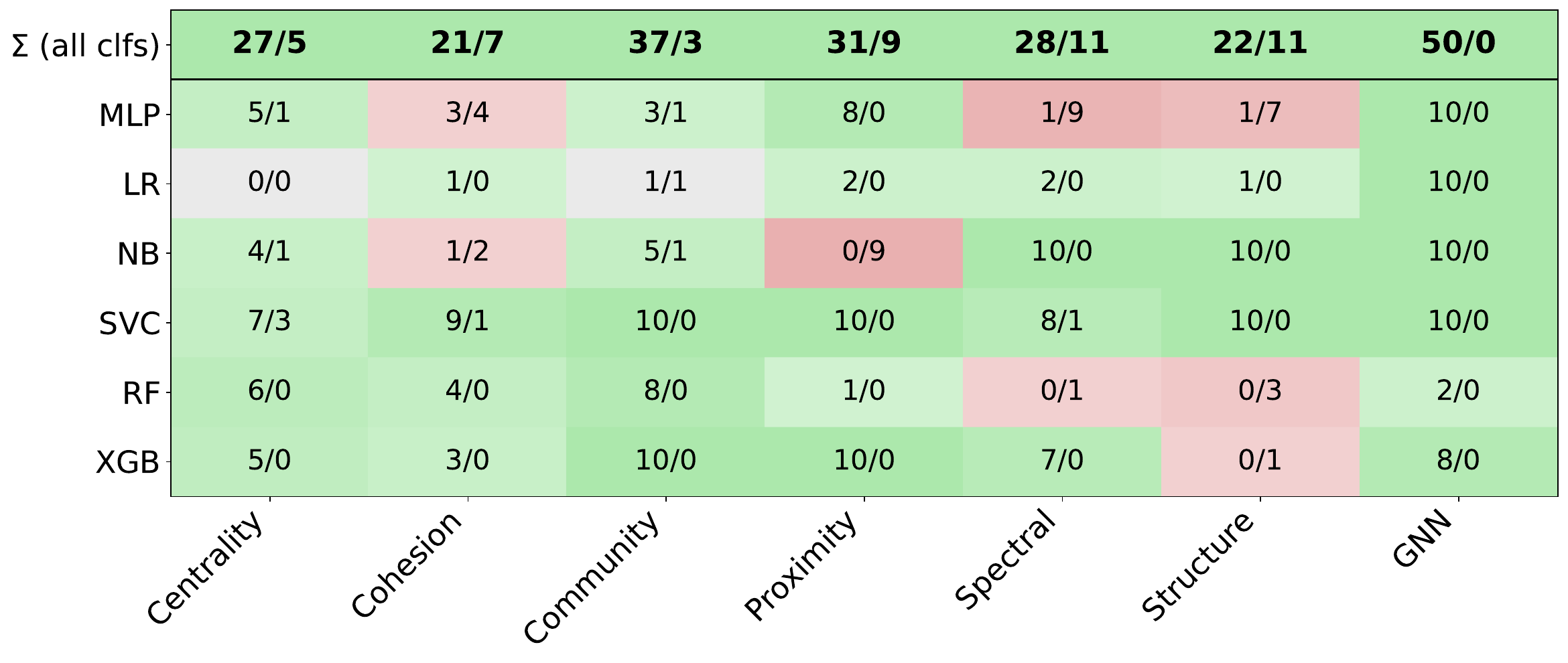}
  \caption{Aggregated McNemar test outcomes for graph signal categories across classifiers.
  Each cell represents the balance of statistically significant improvements ($p \leq 0.05$)
  versus degradations. Darker green shades indicate a higher frequency of significant
  performance gains over the transaction-only baseline.}
  \label{fig:mcnemar_category_heatmap}
\end{figure}

Figure~\ref{fig:mcnemar_category_heatmap} visualizes the aggregated McNemar test outcomes at the graph signal category level, summarizing the number of statistically significant improvements versus degradations across classifiers. The top row summarizes net significance patterns across all classifiers, revealing that significant improvements dominate significant degradations for all examined graph categories, with GNN-derived representations, proximity-based embeddings, and community-based signals exhibiting the strongest and most consistent dominance of significant improvements over degradations.

Overall, these results confirm that the observed performance gains from graph-derived signals are not only numerically meaningful but also statistically reliable under paired, run-level significance testing.

Appendix~\ref{app:mcnemar-individual-signal} provides a fine-grained, per-signal breakdown of McNemar test outcomes across classifiers and runs on a signal level.

\subsection{Robustness under Graph Perturbations}
\label{subsec:robustness-graph-perturbations}
This section reports the results of the edge dropping robustness experiments introduced in Section~\ref{subsec:robustness-experiments}. All graph-derived signals are computed on randomly sparsified transaction graphs with 25 \% and 50 \% edge removal, while all other aspects of the experimental setup remain unchanged. To ensure comparability across perturbation levels, all classifiers are evaluated using the previously determined optimal hyperparameters.

\begin{figure}[H]
  \centering
  \includegraphics[width=\linewidth]{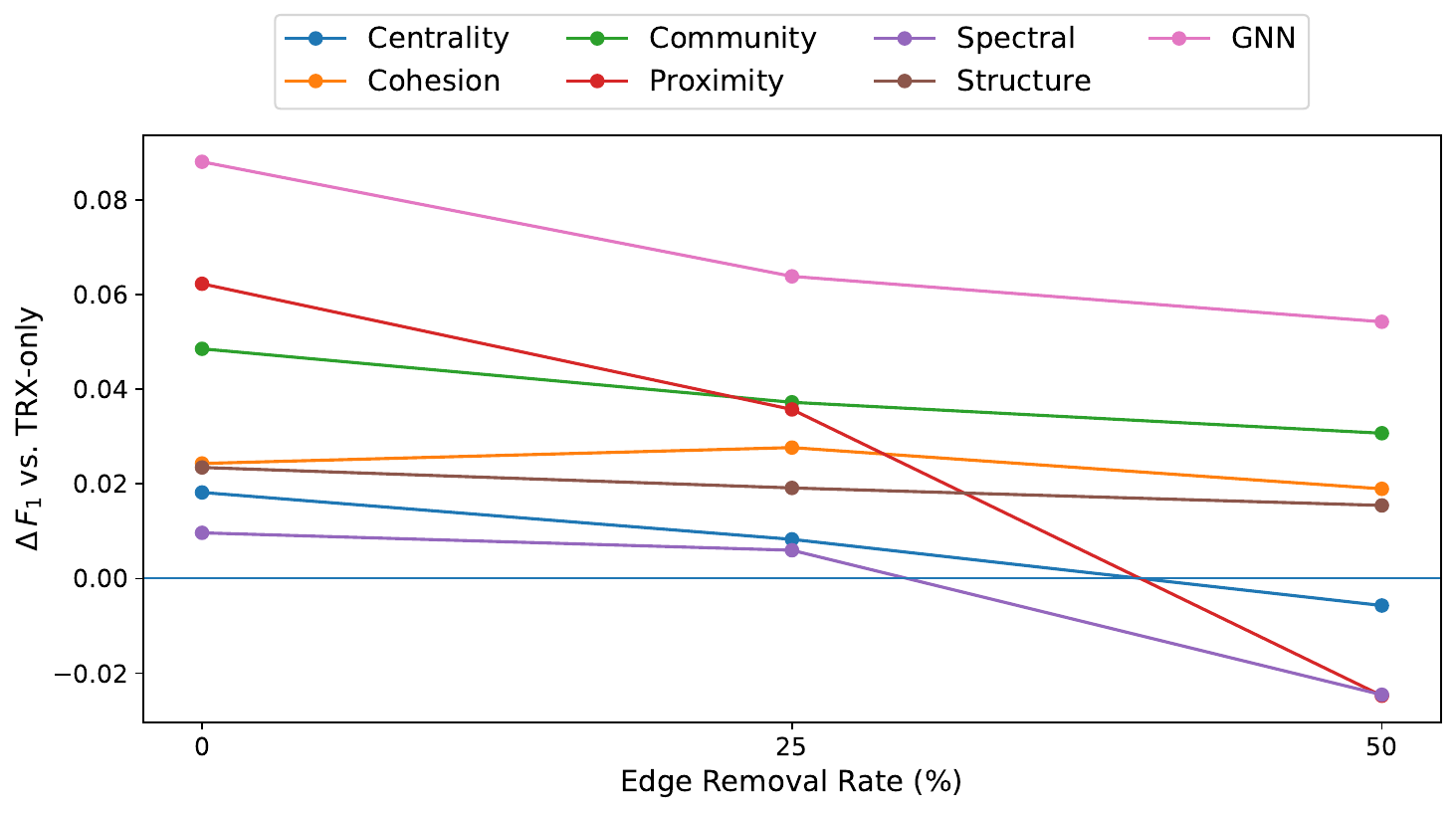}
  \caption{Average $\Delta F_1$ relative to the transaction-only baseline across graph signal
  categories under increasing edge removal.}
  \label{fig:avg_delta_f1_edge_removal}
\end{figure}

Figure~\ref{fig:avg_delta_f1_edge_removal} illustrates the average $\Delta F_1$ relative to the transaction-only baseline across graph signal categories as a function of increasing edge removal aggregated across classifiers. Additional details are reported in Appendix~\ref{app:robustness-graph-perturbations}, where detailed result matrices for each edge removal level are provided in Figure~\ref{fig:signal_classifier_matrix_25} and Figure~\ref{fig:signal_classifier_matrix_50}, and classifier-specific robustness trends are illustrated in Figure~\ref{fig:robustness_trends_edge_removal}.

As illustrated in Figure~\ref{fig:avg_delta_f1_edge_removal}, increasing edge removal leads to distinct degradation profiles across graph signal categories, indicating that robustness under structural perturbation varies substantially across different types of graph-derived signals.

Proximity-based signals exhibit the steepest performance degradation under edge removal, with average $F_1$-score improvements declining rapidly as the graph becomes sparser. This behavior is consistent with the reliance of proximity-based representations on multi-hop neighborhood structure, which is particularly sensitive to random edge deletion.

Community-based signals show a more gradual but consistent decrease in performance, suggesting moderate robustness under structural degradation. While these signals remain beneficial under mild perturbations, their effectiveness diminishes as community structure becomes increasingly fragmented.

Cohesion-based and centrality-based signals display relatively mild degradation on average; however, their robustness is not uniform across classifiers (see Figure~\ref{fig:robustness_trends_edge_removal} in Appendix~\ref{app:robustness-graph-perturbations}). For some downstream models, these signals remain stable under moderate edge removal, whereas for others their contribution deteriorates, indicating classifier-dependent sensitivity.

Structural embeddings exhibit heterogeneous robustness behavior. While certain classifiers retain stable or only mildly degraded performance under edge removal, others show notable declines, highlighting that robustness for this category depends strongly on the interaction between the embedding method and the downstream classifier (see Figure~\ref{fig:robustness_trends_edge_removal} in Appendix~\ref{app:robustness-graph-perturbations}).

In contrast, GNN-derived representations demonstrate the strongest overall robustness across perturbation levels. Although performance declines with increasing edge removal, GNN-based signals consistently retain positive improvements relative to the transaction-only baseline, even under substantial graph sparsification.

\textbf{Summary}

Overall, these results show that robustness under graph perturbations is highly signal- and classifier-dependent. Categories exhibiting similar average performance under the full graph can differ markedly in their sensitivity to structural degradation, underscoring the importance of evaluating robustness alongside effect size when assessing the practical utility of graph-derived signals.

\section{Discussion}
\label{sec:discussion}

The results presented in Section~\ref{sec:experimental-results} provide the empirical basis for addressing the research questions posed in the Introduction. In this section, we synthesize these findings to answer the research questions by relating performance, robustness, and interpretability patterns to different categories of graph-derived signals in the considered application setting.

Specifically, this discussion examines whether graph-derived signals provide statistically reliable performance improvements over transaction-only baselines and how these effects
vary across signal categories (RQ1), how robust the observed performance gains are under controlled graph perturbations (RQ2), and to what extent a taxonomy-driven organization of graph signals supports interpretable, category-level insights into characteristic fraud-related patterns (RQ3).

The experimental results presented in Section~\ref{sec:experimental-results} are the outcome of a deliberately controlled and statistically grounded evaluation protocol. By flexibly combining taxonomy-driven graph signal integration, automated hyperparameter optimization, multi-seed evaluation, and formal significance testing, the proposed evaluation protocol enables reliable, comparable, and interpretable assessment of graph-derived signals for tabular learning.

This study set out to systematically evaluate whether, and under what conditions, graph-derived signals provide measurable, statistically reliable, and robust performance benefits for graph-augmented tabular machine learning. Using fraud detection on transaction networks as a case study, the results provide converging evidence that incorporating relational information yields consistent and statistically significant improvements over transaction-only baselines. At the same time, they reveal that the magnitude, stability, and robustness of these gains depend strongly on the type of graph-derived signal, the downstream classifier, and the quality of the underlying graph structure.

\subsection{Effectiveness of Graph-Derived Signals}
\label{subsec:effectiveness-graph-signals}
Across classifiers, graph signal categories, and individual graph signals, the integration of graph-derived signals leads to consistent and statistically reliable improvements over the transaction-only baseline, with gains observed across the large majority of evaluated classifier–signal combinations rather than being confined to isolated configurations. This indicates that relational information captured from the transaction graph provides complementary predictive power beyond node-local transaction features.

From a qualitative perspective, the effectiveness of graph-derived signals is strongly influenced by the type of relational information they encode. Signals that encode local relational proximity and neighborhood context emerge as particularly beneficial in this setting. This observation aligns with the intuition that illicit behavior in transaction networks often propagates through localized neighborhoods, making proximity-aware representations well suited for fraud detection tasks. In contrast, more global or frequency-based representations provide only limited additional benefit, suggesting that not all forms of structural information are equally informative for this application domain.

Although proximity-based and community-based signals are treated as distinct categories within the proposed taxonomy, both capture aspects of relational closeness at different levels of abstraction. The stronger performance of proximity-based representations suggests that modeling relational proximity in a continuous and fine-grained manner can convey richer information than discrete group assignments in the evaluated fraud detection setting. This highlights the importance of considering not only which structural properties are encoded, but also how they are represented when assessing the utility of graph-derived signals.

Taken together, these findings underscore that the effectiveness of graph-based augmentation is closely tied to the alignment between signal type, representation granularity, and the relational characteristics of the underlying task.

\subsection{Classifier-Specific Interactions and Model Capacity}
\label{subsec:classifier-interactions-capacity}
The effectiveness of graph signals varies substantially across classifiers, reflecting differences in model capacity and inductive bias. Tree-based ensemble models, particularly Random Forests and XGBoost, benefit most consistently from graph-based augmentation. These models are well-suited to exploit heterogeneous, potentially redundant feature sets and exhibit stable improvements across nearly all graph signal categories. These findings align with broader evidence that, in tabular learning settings, model performance is often driven more by feature quality than by architectural complexity~\cite{Grinsztajn.2022}.

Support Vector Classifiers show pronounced relative improvements over the transaction-only baseline across many graph signal configurations. However, these gains are accompanied by increased variability across random seeds, reflecting the known sensitivity of SVM-based models to data splits, feature scaling, and hyperparameter settings. While graph signals can substantially enhance SVC performance, these results underscore the importance of careful model tuning and stability analysis when using margin-based classifiers in graph-augmented settings.

Linear models and shallow neural networks display more moderate but robust improvements, suggesting that even comparatively simple classifiers can benefit from relational information when graph-derived signals are appropriately constructed. In contrast, Naive Bayes classifiers exhibit minimal sensitivity to graph augmentation, indicating limited compatibility between conditional independence assumptions and graph-derived signal representations.

\subsection{Stability, Variability, and Robustness}
\label{subsec:stability-variability-robustness}
Beyond average performance gains, the stability of graph-augmented models across random seeds is a critical consideration. The results demonstrate that standard deviations of $F_1$-scores remain low across most classifiers and graph signal categories, typically on the order of a few thousandths. In many cases, graph augmentation even reduces run-to-run variability relative to the transaction-only baseline, indicating a stabilizing effect on model behavior.

Importantly, graph signal categories exhibiting similar average performance under the full graph can differ markedly in their robustness to structural perturbations, highlighting that average effect size alone is insufficient for assessing practical utility without explicit robustness evaluation.

From a practical perspective, these findings imply that the choice of graph-derived signals should be informed not only by average performance under ideal graph conditions, but also by the expected level of noise, incompleteness, or uncertainty in the underlying graph structure. Robustness patterns vary substantially across graph signal categories and perturbation levels, underscoring the importance of signal-type-aware robustness evaluation when selecting graph-derived indicators for downstream tasks.

At the same time, observed performance gains consistently exceed the corresponding variability across runs, indicating that the improvements reflect systematic effects of incorporating relational information.

Finally, the results indicate that the benefit of graph-derived signals is not entirely classifier-invariant, suggesting that signal selection and classifier choice should be considered jointly when designing graph-augmented tabular learning pipelines.

\subsection{Statistical Significance of Performance Improvements}
\label{subsec:statistical-significance-discussion}
Paired McNemar significance tests provide a statistical assessment of performance differences between graph-augmented models and transaction-only baselines. Across a large number of paired comparisons arising from the multi-classifier, multi-signal-group, and multi-seed evaluation setup introduced in Section~\ref{sec:methodology}, a substantial majority of comparisons yield statistically significant improvements over the transaction-only baseline. This pattern persists when results are aggregated at both the graph signal category level and the individual signal level, indicating that beneficial effects are systematic rather than incidental.

Proximity-based and GNN-derived representations exhibit particularly strong and consistent significance patterns, while spectral features show mixed outcomes, aligning with their limited average performance improvements. Importantly, significant degradations are rare overall and occur primarily in isolated classifier–signal combinations rather than as systematic trends.

Taken together, these results demonstrate that graph-derived signals not only improve predictive performance numerically, but do so in a statistically robust and reproducible manner.

\subsection{Interpretation and Methodological Implications}
\label{subsec:interpretation-methodological-implications}
The findings of this study demonstrate that incorporating graph-derived signals into tabular learning pipelines yields statistically significant and reproducible performance improvements in the context of cryptocurrency transaction fraud detection.

From a methodological perspective, the proposed evaluation protocol enables a systematic and taxonomy-driven integration of diverse graph-derived signals into tabular learning pipelines. By supporting flexible signal grouping, controlled hyperparameter optimization, and formal statistical evaluation, it allows practitioners to identify which types of graph signals yield statistically significant and stable benefits for a given downstream task. In particular, the results highlight the importance of selecting graph signal types that align with the structural characteristics of the target problem, as both performance impact and robustness under structural perturbations vary substantially across signal categories.

Moreover, the inclusion of controlled structural perturbations provides a practical mechanism to assess how different graph signal categories respond to incomplete or degraded graph structure. The observed heterogeneity in degradation patterns underscores that robustness is highly signal-type dependent and should be considered explicitly during signal selection.

In the evaluated cryptocurrency transaction fraud setting, graph neural network–based representations achieve the strongest overall performance gains. Notably, GNNs are the only class of methods considered in this study that explicitly incorporate feature information from neighboring nodes during representation learning. While multiple factors may contribute to their effectiveness, this observation highlights the potential relevance of local relational context in this application domain.

Beyond GNN-based representations, proximity-based and community-oriented graph signals also exhibit strong and consistent performance improvements, suggesting that relational closeness and mesoscopic grouping information are particularly informative. In contrast, purely structural or spectral representations show more heterogeneous effects. Importantly, these observations are specific to the analyzed transaction network and should not be interpreted as a general ranking of graph signal types across domains.

Beyond predictive performance, the taxonomy-driven evaluation of graph signal categories provides a basis for pattern-oriented analysis of downstream tasks. By examining which types of graph-derived signals yield consistent and robust improvements, the evaluation protocol supports high-level interpretation of which structural aspects of the graph are informative for the task, such as fraud detection in transaction networks.

\subsection{Practical Implications and Business Value}
\label{subsec:practical-implications-business-value}
The empirical results of this study have direct practical implications for fraud detection systems operating under severe class imbalance and regulatory constraints. They demonstrate that explicitly extracted graph-derived signals can yield consistent and statistically significant improvements in fraud detection performance when integrated into classical tabular machine learning pipelines, improving both precision and recall, which are critical performance dimensions in financial crime detection.

From a practical perspective, the primary contribution of the proposed evaluation protocol lies not in prescribing a fixed ranking of graph signal types for a given application task, but in providing a systematic and statistically grounded procedure to assess their utility for a given downstream task.

As demonstrated in the presented fraud detection case study, the relative effectiveness, stability, and robustness of graph-derived signals vary substantially across signal categories and classifiers. Importantly, these patterns are task-dependent and should not be assumed to generalize across application domains. The proposed evaluation protocol enables practitioners to identify, under controlled experimental conditions, which types of graph-derived signals provide statistically significant and robust benefits for their specific task, data characteristics, and operational constraints. In practical terms, this implies that signal selection should account for known data quality constraints of the application. For example, in settings where transaction graphs are expected to be incomplete or partially observed—such as when access to certain transaction channels is limited—the results of our case study suggest that placing greater emphasis on more coarse-grained or role-based graph signal categories may yield more stable performance than relying primarily on proximity-based signals, which exhibited pronounced sensitivity to graph perturbations.

In regulated environments, interpretability and auditability are often as important as predictive performance. Several graph signal categories that perform particularly well in the evaluated setting, such as community-based indicators are inherently interpretable and provide transparent representations of relational structure. This facilitates signal-level transparency and model auditability without relying on opaque end-to-end graph neural architectures.

Finally, the robustness experiments in our case study highlight that different graph signal categories exhibit markedly different sensitivity to incomplete or degraded graph structure. In the evaluated fraud detection setting, this allows practitioners to identify which types of graph-derived signals remain informative when transaction graphs are sparse, partially observed, or subject to data loss. The proposed evaluation protocol therefore provides a principled mechanism to assess, for a given downstream task, which graph-derived signals retain predictive utility under incomplete or degraded graph structure.

\subsection{Limitations and Future Directions}
\label{subsec:limitations-future-directions}

While the proposed evaluation protocol and taxonomy-driven analysis are designed to be generic and applicable to a broad range of graph-augmented tabular learning tasks, the empirical results presented in this study are obtained from a single large-scale cryptocurrency fraud detection benchmark. Consequently, the quantitative performance gains, robustness patterns, and relative effectiveness of different graph signal categories should be interpreted in the context of this specific application setting.

At the same time, the Elliptic dataset constitutes a challenging and representative real-world benchmark, characterized by strong relational dependencies, severe class imbalance, and noisy graph structure. These properties make it well suited for stress-testing the statistical reliability and robustness of graph-derived signals. The observed patterns—such as the strong effectiveness of proximity-based signals and their pronounced sensitivity to graph perturbations—are therefore indicative of how different types of graph-derived signals behave under realistic conditions, rather than being benchmark-specific performance claims.

Importantly, the primary contribution of this work lies not in the absolute performance levels achieved on this dataset, but in the systematic, statistically grounded evaluation protocol and taxonomy-driven perspective. These components are directly transferable to other application domains and datasets, enabling comparable analyses of graph-derived signals beyond the specific case study considered here.

The set of graph-derived signals evaluated within each category is necessarily non-exhaustive. Although representative signal types are selected to cover a broad spectrum of structural information, additional graph signals and alternative formulations may further enrich the analysis. Extending the set of signal representatives per category may provide finer-grained insights into category-internal variability.

Several design choices were made to prioritize comparability and statistical rigor, which also introduce constraints; graph signals are computed deterministically and held fixed across random seeds, enabling clean paired comparisons across classifiers and evaluation runs. However, this setup does not explicitly capture additional uncertainty arising from alternative graph construction procedures or stochastic graph signal generation. Similarly, while hyperparameter optimization is applied consistently across models, broader hyperparameter spaces, additional optimization trials, or alternative optimization objectives may further improve absolute performance.

Despite extensive measures to mitigate variability, including automated hyperparameter optimization, multiple random seeds, diverse classifier families, formal significance testing, and trimmed performance aggregation, residual stochastic effects and dataset-specific biases may still contribute to the observed results, as in all empirical machine learning studies.

Finally, we note that the proposed taxonomy constitutes an abstraction of a continuous and heterogeneous design space. While the categories capture distinct types of structural information, boundaries between categories are not always strict, and certain graph signals may exhibit characteristics of multiple groups. The taxonomy is therefore intended as a practical organizing scheme rather than a rigid classification.

Future work may extend the proposed evaluation protocol along several directions, including evaluating additional graph signal representatives and classifier families, incorporating stochastic or task-adaptive graph signal generation, extending robustness analysis to dynamic or temporally evolving graphs, and exploring alternative evaluation protocols beyond binary fraud detection. More generally, the proposed evaluation protocol provides a foundation for systematic and extensible benchmarking of graph-derived signals under controlled experimental conditions.

\section{Conclusion}
\label{sec:conclusion}

This paper introduces a systematic, taxonomy-driven evaluation protocol for assessing graph-derived signals in tabular machine learning. The proposed taxonomy provides the conceptual framework that enables a structured and interpretable comparison across fundamentally different types of relational information. The protocol combines multi-seed statistical evaluation, formal significance testing, and robustness analysis under graph perturbations to enable fair and reliable comparisons.

Applied to a large-scale cryptocurrency fraud detection case study, the taxonomy-guided protocol yields three key insights. First, while graph-derived signals consistently provide statistically significant improvements over tabular baselines, their effectiveness varies substantially across the distinct categories defined by the taxonomy. Proximity-based and GNN-derived signals emerge as particularly impactful, suggesting that localized relational patterns are highly discriminative for fraud detection. Second, robustness under structural perturbations is highly signal-dependent, revealing distinct degradation profiles per category. For instance, while proximity-based signals perform well on the complete graph, they degrade rapidly under edge removal, making them less suitable for applications with incomplete or noisy transaction data. Third, the taxonomy-driven analysis provides interpretable insights by linking these highly discriminative signal categories to characteristic structural fraud patterns, such as the reliance on localized transaction neighborhoods.

Overall, this work demonstrates the importance of assessing graph-derived signals using statistically grounded evaluation criteria beyond average performance comparisons. Accordingly, meaningful evaluation requires jointly considering effect size, statistical reliability, and robustness under structural uncertainty, which capture complementary aspects of signal utility and should not be conflated when assessing practical applicability. The proposed taxonomy further facilitates interpretable insights in applied domains. Using fraud detection as a representative case study, we show how the proposed evaluation protocol enables principled and reproducible assessment of the conditions under which graph-derived signals provide measurable value for tabular machine learning models.

Methodologically, the evaluation protocol provides an application-agnostic methodology for the informed selection of graph signals based on these complementary criteria. It can be directly applied to other domains to determine when and how graph-derived information reliably enhances tabular learning pipelines.

\vspace{6pt} 

\section*{Funding}
This research received no external funding.

\section*{Data Availability}
The data and source code supporting the findings of this study are publicly available in the GitHub repository at \url{https://github.com/graph-eval/graph-eval-protocol}. The experimental source code and result artifacts are also available on Zenodo at \url{https://doi.org/10.5281/zenodo.18351526}.

\section*{Acknowledgments}
The authors would like to thank Prof. Dr. Mª Mercedes Carmona Martínez and Dr. Vinny Flaviana Hyunanda from Universidad Católica San Antonio de Murcia (UCAM), as well as Claudia Schmitz from FOM University of Applied Sciences, for their kind administrative support throughout the doctoral process.


\section*{Conflicts of Interest}
The authors declare no conflicts of interest.

\appendix
\section{Mean F1-score per individual graph signal}
\label{app:mean-f1}

\begin{figure}[H]
  \centering
  \includegraphics[width=\linewidth]{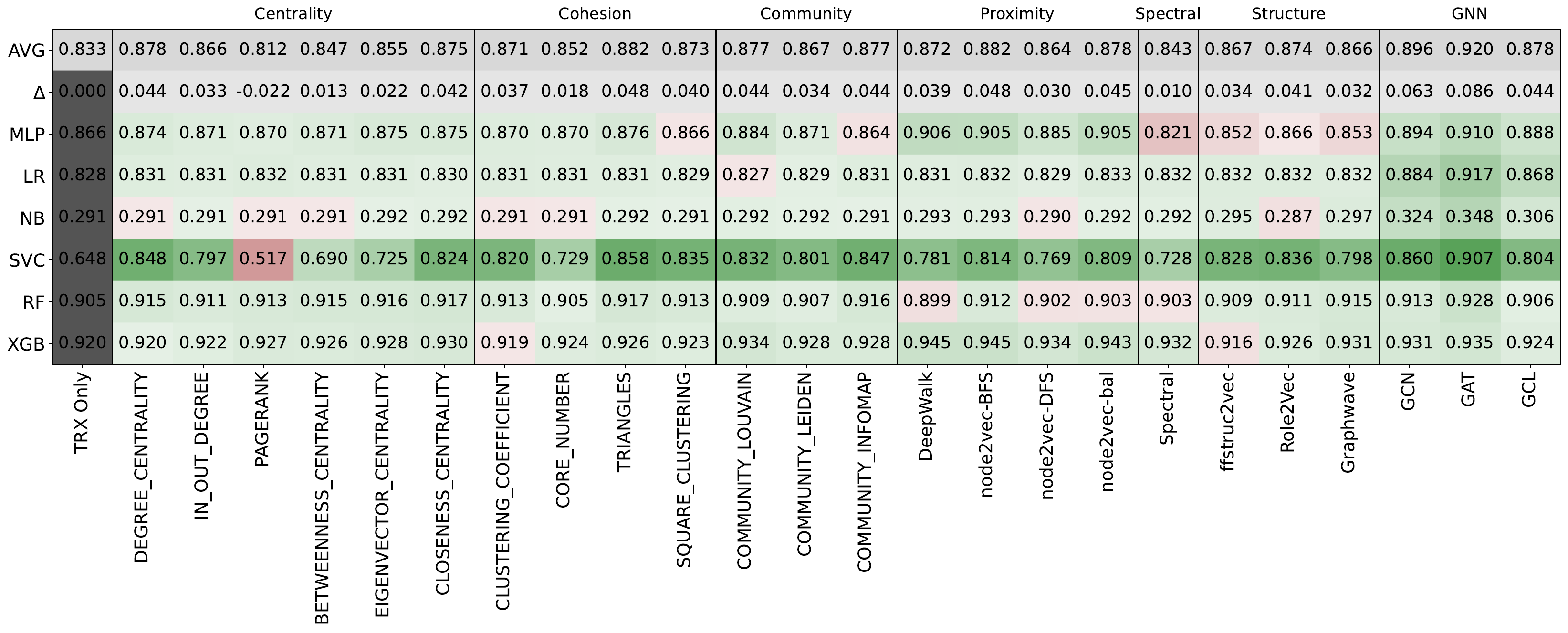}
  \caption{Mean $F_1$-scores across classifiers and graph signals.
  Cell values report the mean $F_1$-score aggregated over the trimmed middle eight runs.
  Color intensity indicates relative performance differences with respect to the
  transaction-only baseline (green = improvement, red = degradation).}
  \label{fig:mean_f1_matrix}
\end{figure}

We analyze the distribution of performance changes across all classifier–graph signal combinations, as illustrated in Figure~\ref{fig:mean_f1_matrix}. Out of 144 evaluated combinations (6 classifiers × 24 graph signals), 123 (85.4\%) yield an improvement in $F_1$-score relative to the transaction-only baseline, while only 21 (14.6\%) result in a decrease. Averaged across all graph signal combinations, graph augmentation leads to a mean $F_1$-score increase of +0.031, indicating that performance gains are not confined to isolated configurations but occur consistently across models and signal types. The corresponding variability of these performance estimates across random seeds is reported in Appendix~\ref{app:std-f1}, allowing effect magnitude and stability to be assessed jointly.

\section{Standard deviation per individual graph signal}
\label{app:std-f1}

\begin{figure}[H]
  \centering
  \includegraphics[width=\linewidth]{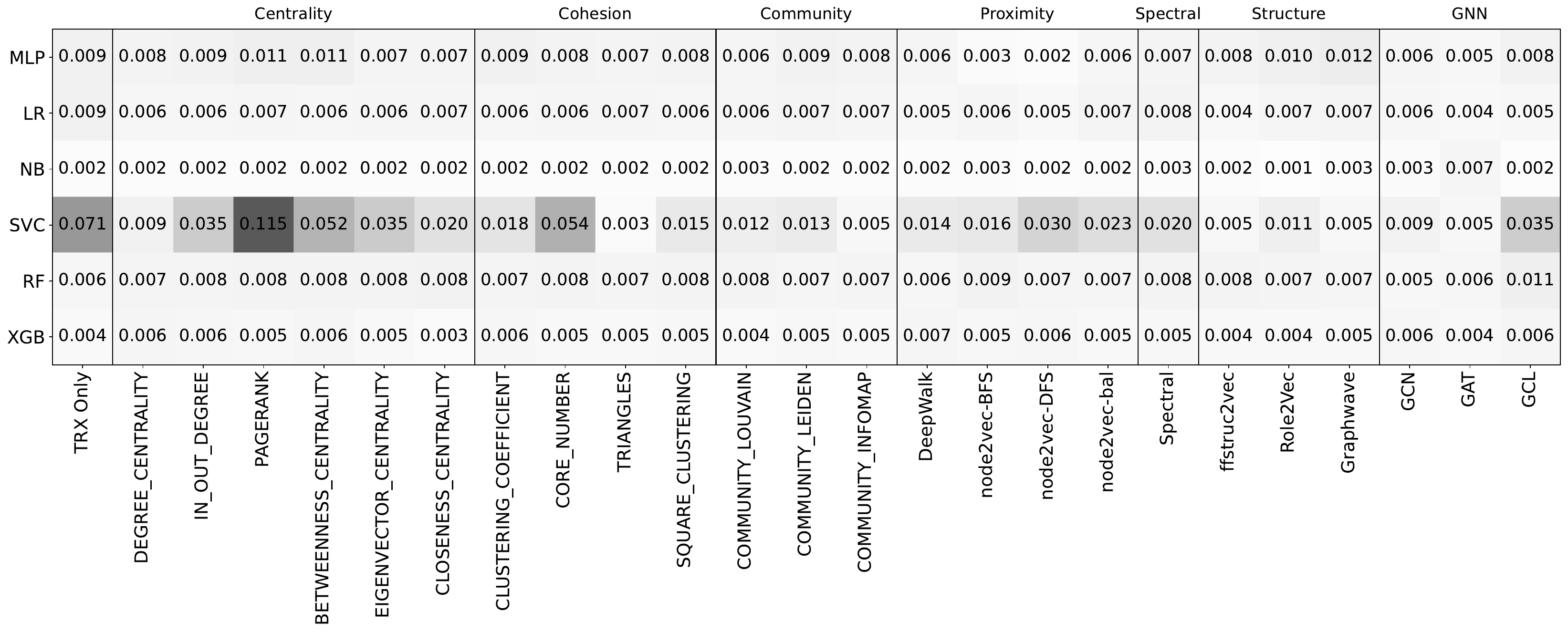}
  \caption{Standard deviation of $F_1$-scores across random seeds. Values report the standard deviation of $F_1$-scores over the trimmed middle eight runs. Darker shading indicates higher variability. Graph signals are grouped by category.}
  \label{fig:std_f1_signal_matrix}
\end{figure}

This appendix reports the standard deviation of $F_1$-scores across random seeds for each classifier–graph signal combination. Together with the mean performance values in Appendix~\ref{app:mean-f1}, these results allow the stability of graph-derived signals to be assessed alongside their average effect size.

\section{Statistical Significance Analysis - McNemar per individual signal}
\label{app:mcnemar-individual-signal}
\begin{figure}[H]
  \centering
  \includegraphics[width=\linewidth]{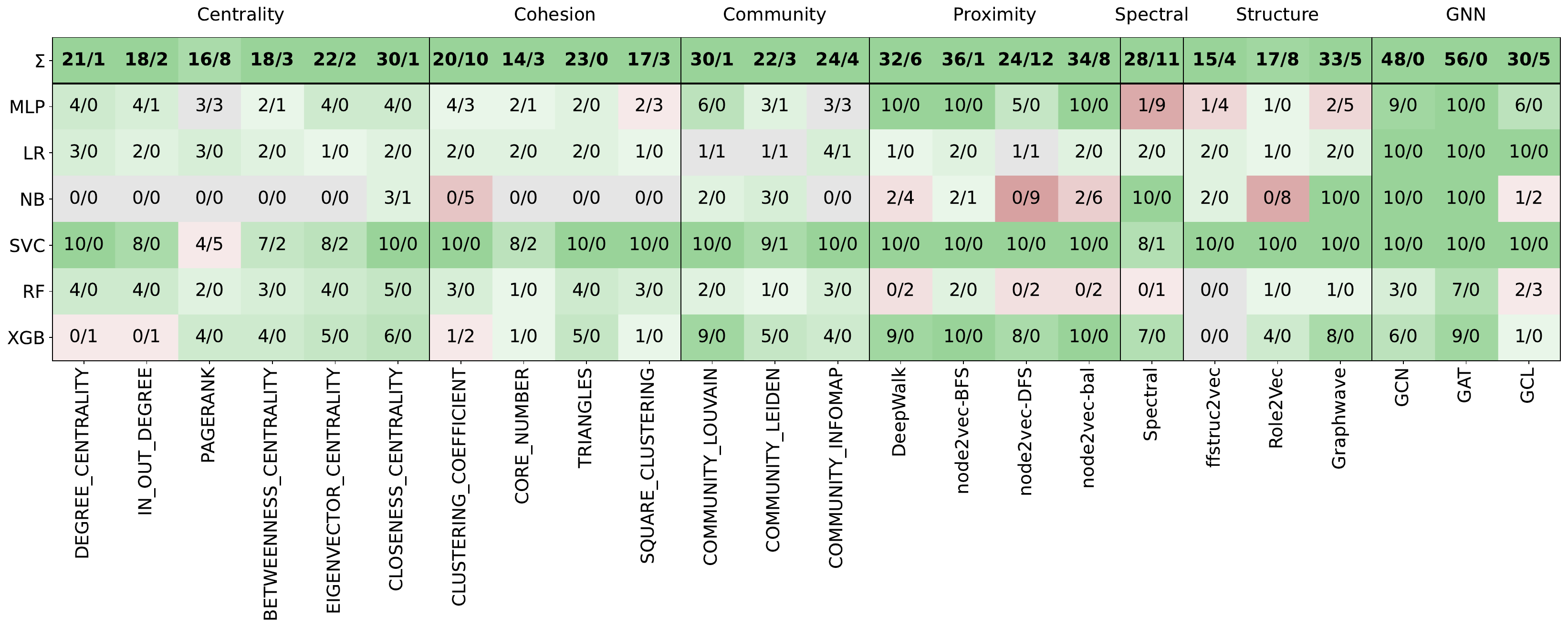}
  \caption{Signal-level aggregation of McNemar test outcomes across classifiers and runs ($p \leq 0.05$). 
  Cells show the number of statistically significant improvements versus degradations (\#better / \#worse) 
  for each classifier--graph signal combination, aggregated over ten random seeds. 
  Color intensity reflects the net balance between improvements and degradations.}
  \label{fig:mcnemar_signal_level}
\end{figure}

Across all evaluated configurations, a total of 1,440 paired comparisons were conducted (6 classifiers $\times$ 24 graph signals $\times$ 10 random seeds). Out of these, 628 comparisons (43.6\%) show a statistically significant improvement over the transaction-only baseline at a significance level of $p \leq 0.05$, whereas only 101 comparisons (7.0\%) exhibit a significant degradation. This asymmetry indicates a strong overall bias toward beneficial effects of graph-based augmentation.

\section{Robustness Analysis under Graph Perturbations}
\label{app:robustness-graph-perturbations}
\begin{figure}[H]
  \centering
  \includegraphics[width=\linewidth]{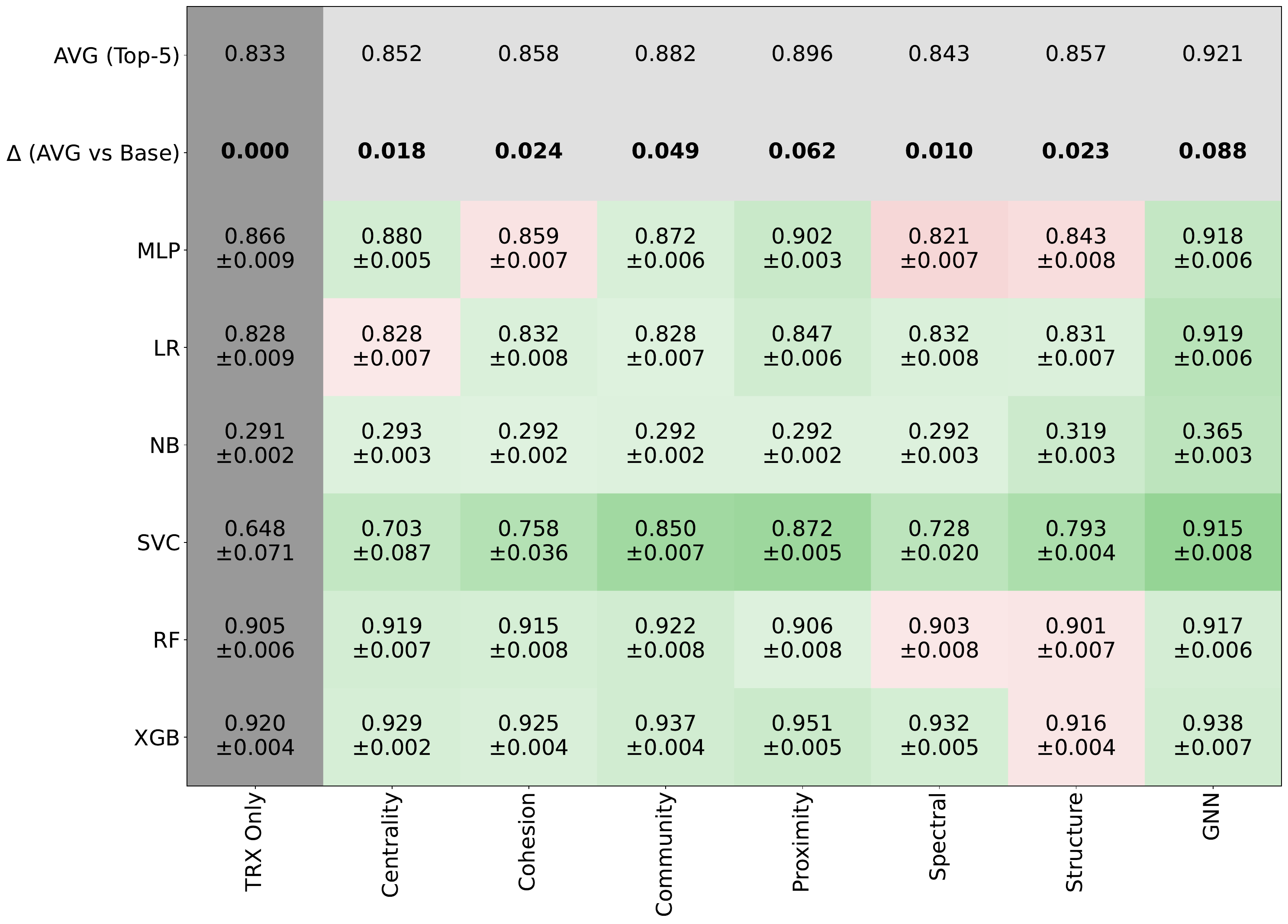}
  \caption{Performance comparison under structural graph perturbations with 0\% edge removal
(transaction-only reference). Cell values report mean $F_1$-scores with standard deviations
across random seeds.}
  \label{fig:signal_classifier_matrix_0}
\end{figure}

\begin{figure}[H]
  \centering
  \includegraphics[width=\linewidth]{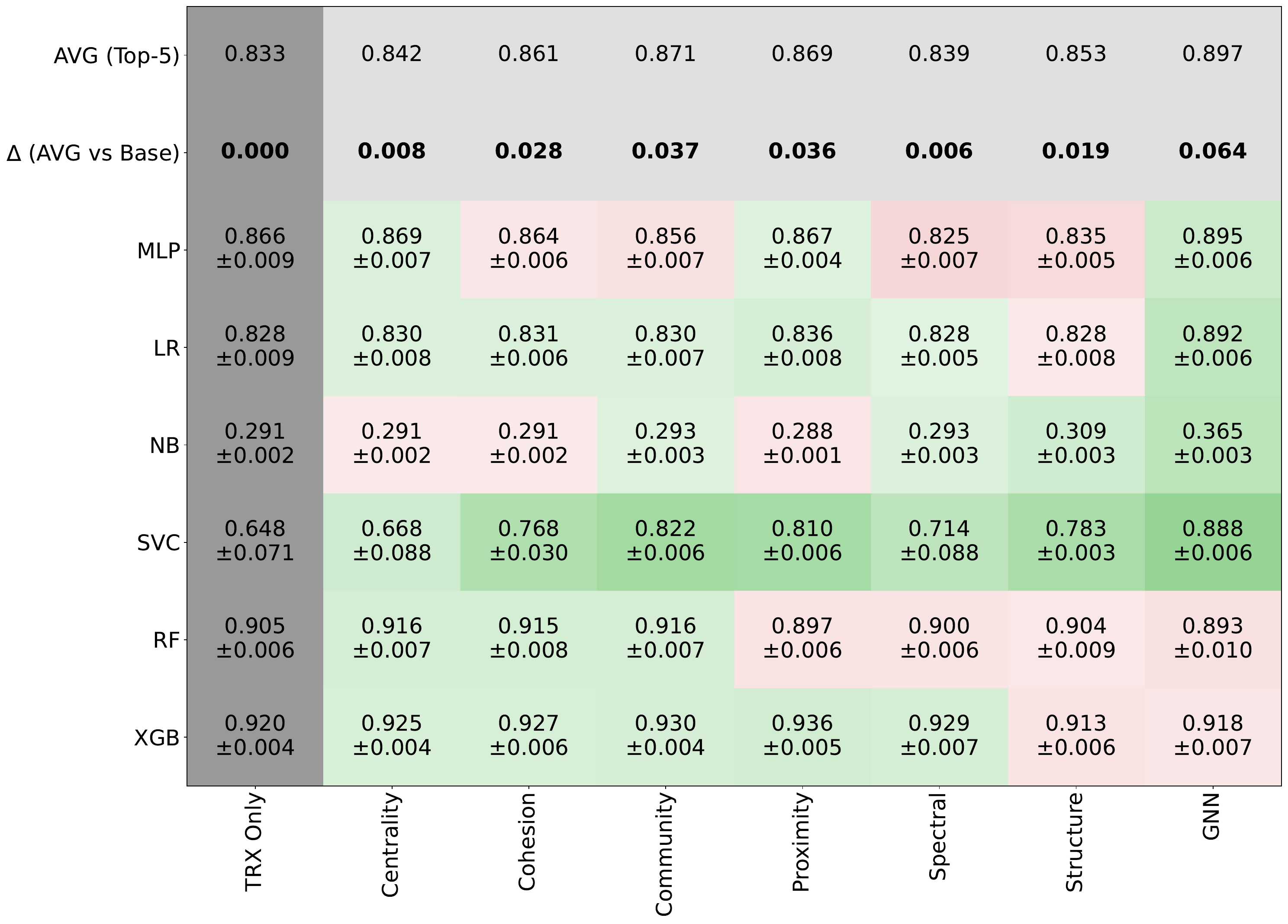}
  \caption{Performance comparison under moderate structural degradation with 25 \% random edge removal.}
  \label{fig:signal_classifier_matrix_25}
\end{figure}

\begin{figure}[H]
  \centering
  \includegraphics[width=\linewidth]{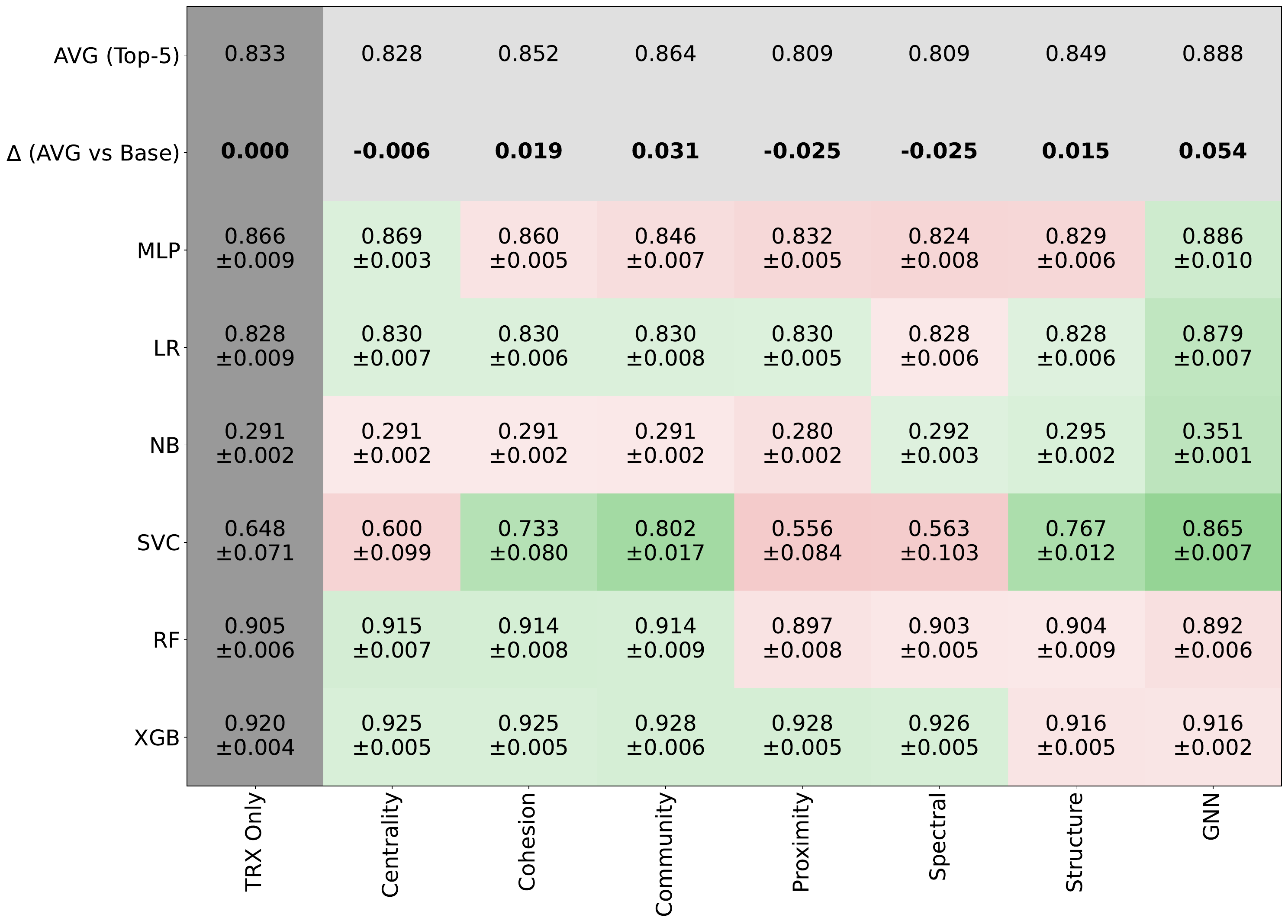}
  \caption{Performance comparison under stronger structural degradation with 50 \% random edge removal.}
  \label{fig:signal_classifier_matrix_50}
\end{figure}

\begin{figure}[H]
  \centering
  \includegraphics[width=\linewidth]{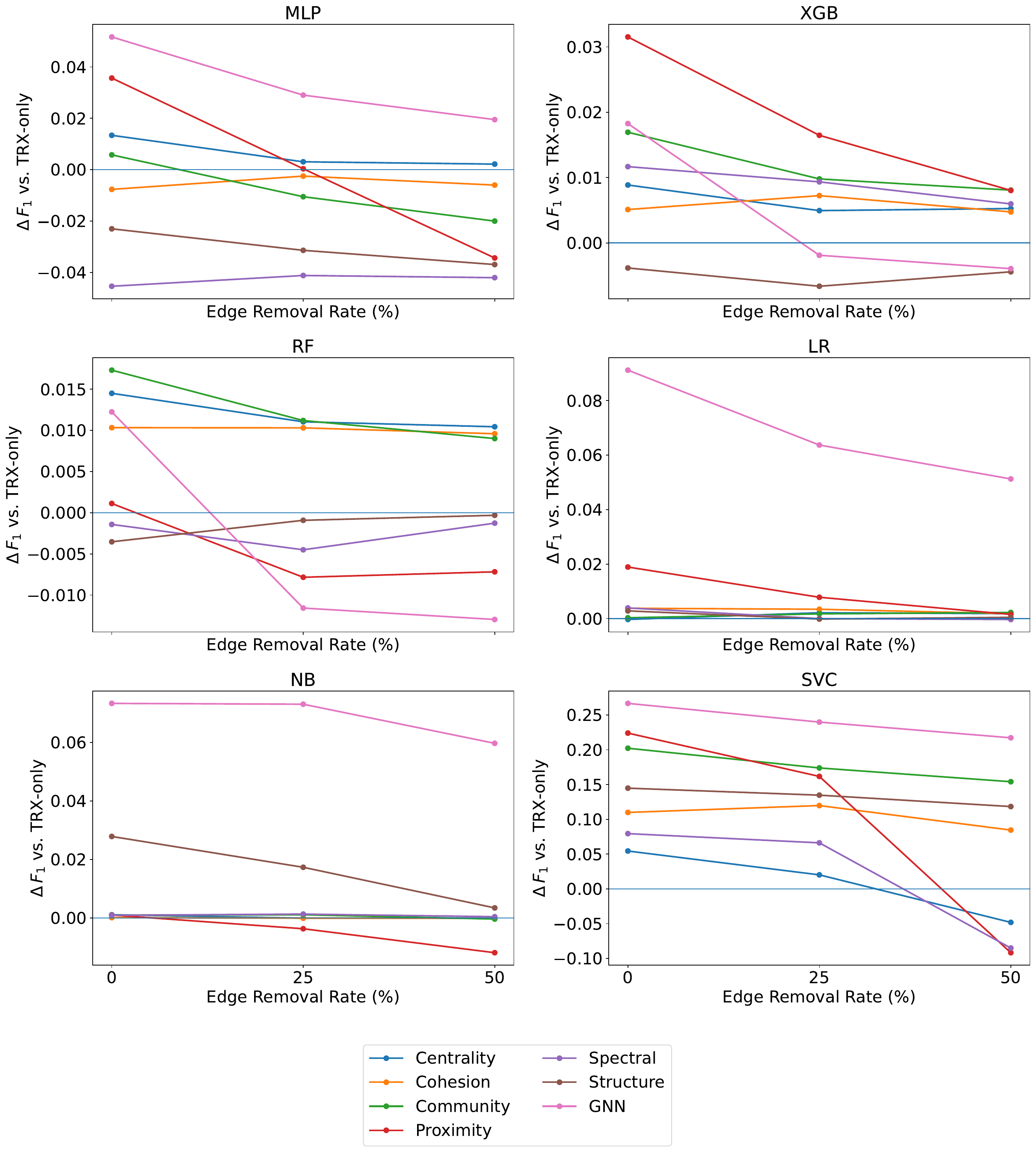}
  \caption{Classifier-specific robustness trends illustrating how average $\Delta F_1$
  improvements relative to the transaction-only baseline evolve under increasing levels
  of random edge removal (0\%, 25\%, 50\%) for different graph signal categories.}
  \label{fig:robustness_trends_edge_removal}
\end{figure}


\section{Classification Hyperparameters}
\label{app:ClassificationHyperparameters}

This appendix reports the complete hyperparameter search spaces used for all evaluated classifiers.
All hyperparameters listed here are optimized via Bayesian optimization using Hyperopt with a fixed
budget of 50 trials per configuration. Parameters not listed are kept at library default values.

\begin{table}[H]
\caption{Hyperparameter search spaces for all evaluated classifiers.
Only the parameters listed are optimized via Bayesian optimization (Hyperopt)
with a fixed trial budget; all remaining parameters are kept at library default values.}
\label{tab:S1_classifier_hyperparameters}
\centering
\small
\begin{tabular}{p{0.30\linewidth} p{0.22\linewidth} p{0.43\linewidth}}
\toprule
\textbf{Model} & \textbf{Parameter} & \textbf{Search Space} \\
\midrule
Logistic Regression & penalty & $\{l2,\ \texttt{None}\}$ \\
 & $C$ & log-uniform $[0.001,\ 10]$ \\
 & solver & $\{\texttt{lbfgs},\ \texttt{saga}\}$ \\
 & balancing\_strategy & $\{\texttt{none},\ \texttt{oversampling},\ \texttt{class\_weights}\}$ \\
 & use\_emb\_pca & $\{\texttt{False},\ \texttt{True}\}$ \\
 & emb\_pca\_n & $\{16,\ 32,\ 48,\ 64,\ 0.95\}$ \\
\midrule
Random Forest & n\_estimators & quniform $[100,\ 600]$, step 25 \\
 & max\_depth & $\{\texttt{None},\ 5,\ 8,\ 12,\ 16,\ 20,\ 30,\ 40\}$ \\
 & min\_samples\_split & quniform $[2,\ 10]$, step 1 \\
 & min\_samples\_leaf & quniform $[1,\ 5]$, step 1 \\
 & max\_features & $\{\texttt{sqrt},\ \texttt{log2},\ \texttt{None}\}$ \\
 & balancing\_strategy & $\{\texttt{none},\ \texttt{oversampling},\ \texttt{class\_weights}\}$ \\
\midrule
SVC (linear kernel) & SVC\_C & log-uniform $[10^{-3},\ 10^{2}]$ \\
\midrule
SVC (RBF kernel) & SVC\_C & log-uniform $[10^{-3},\ 10^{2}]$ \\
 & gamma & log-uniform $[10^{-4},\ 10^{1}]$ \\
\midrule
XGBoost & n\_estimators & quniform $[50,\ 500]$, step 25 \\
 & max\_depth & quniform $[2,\ 8]$, step 1 \\
 & learning\_rate & log-uniform $[0.01,\ 0.3]$ \\
 & subsample & uniform $[0.6,\ 1.0]$ \\
 & colsample\_bytree & uniform $[0.6,\ 1.0]$ \\
 & min\_child\_weight & log-uniform $[10^{-1},\ 10]$ \\
 & gamma & log-uniform $[10^{-3},\ 1]$ \\
 & reg\_alpha & log-uniform $[10^{-6},\ 1]$ \\
 & reg\_lambda & log-uniform $[10^{-3},\ 10]$ \\
 & balancing\_strategy & $\{\texttt{none},\ \texttt{oversampling}\}$ \\
\midrule
MLP & hidden\_layer\_sizes & choice from $(64)$, $(128)$, $(256)$, $(64,32)$, $(128,64)$, $(128,128)$, $(256,128)$ \\
 & alpha & log-uniform $[10^{-6},\ 10^{-2}]$ \\
 & lr\_init & log-uniform $[10^{-4},\ 5\times10^{-2}]$ \\
 & activation & $\{\texttt{relu},\ \texttt{tanh}\}$ \\
 & batch\_size & $\{64,\ 128,\ 256\}$ \\
 & balancing\_strategy & $\{\texttt{none},\ \texttt{oversampling}\}$ \\
\midrule
Naive Bayes & var\_smoothing & log-uniform $[10^{-6},\ 10^{-1}]$ \\
 & balancing\_strategy & $\{\texttt{none},\ \texttt{oversampling},\ \texttt{class\_prior\_}\}$ \\
 & oversample\_ratio & uniform $[0.5,\ 1.0]$ \\
 & scaler & $\{\texttt{standard},\ \texttt{minmax}\}$ \\
\bottomrule
\end{tabular}
\end{table}

\section{Graph Signal Generation Parameters}
\label{app:GraphSignalGenerationParameters}

This appendix summarizes all parameters used for the generation of graph-derived signals.
All graph signals are computed once on the fixed transaction graph and reused across
all evaluation runs to ensure paired comparability.

\begin{table}[H]
\caption{Graph signal generation parameters used across all evaluated signal categories.
All graph-derived signals are computed once on the fixed transaction graph and reused
across all evaluation runs to ensure paired comparability.}
\label{tab:S2_graph_signal_parameters}
\centering
\small
\renewcommand{\arraystretch}{1.5}
\begin{tabular}{p{0.28\linewidth} p{0.67\linewidth}}
\toprule
\textbf{Graph Signal} & \textbf{Parameters} \\
\midrule
PageRank &
Damping factor $\alpha = 0.85$; maximum iterations: 100; convergence tolerance: $10^{-8}$; directed graph \\

Betweenness Centrality &
Approximate computation for large graphs; sampled nodes: $\min(2000, |V|)$; undirected graph \\

Eigenvector Centrality &
Computed per connected component; initialization: uniform; maximum iterations: 1000; convergence tolerance: $10^{-6}$; undirected graph \\

Closeness Centrality &
Computed per connected component; undirected graph \\

Infomap &
Directed graph; unweighted edges \\

Leiden &
Undirected graph; resolution parameter: default \\

Proximity-Based Embeddings &
Number of walks per node: 10; walk length: 20; context window size: 10; embedding dimension: 64 \\

role2vec &
Embedding dimension: 64; number of walks: 10; walk length: 20 \\

GraphWave &
Target embedding dimension: 64; number of diffusion scales: 6; diffusion scale range $\tau \in [10^{-2}, 10^{1}]$ (log-spaced); eigen-decomposition: exact dense for small components, truncated sparse for components $\geq 1000$ nodes; retained eigenpairs: 32; Gaussian random projection \\

ffstruc2vec &
Embedding dimension: 64; number of walks: 10; walk length: 20; number of layers considered: 2; cost function: mean- and variance-based structural distance; layer weights: layers 0--2: 1.0, layers $\geq$3: 0.0 \\
\bottomrule
\end{tabular}
\end{table}

\setlength{\tabcolsep}{4pt}
\renewcommand{\arraystretch}{1.2}

\begin{longtable}{p{0.22\linewidth} p{0.30\linewidth} p{0.43\linewidth}}
\caption{Hyperparameter search space for GNN-based models.
The table reports the ranges and options considered during automated Bayesian
hyperparameter optimization for GCN, GAT, and GCL-based models.
All configurations are evaluated under identical optimization budgets and selection criteria.}
\label{tab:S3_gnn_hyperparameters}\\
\toprule
\textbf{Model} & \textbf{Parameter} & \textbf{Search Space} \\
\midrule
\endfirsthead

\toprule
\textbf{Model} & \textbf{Parameter} & \textbf{Search Space} \\
\midrule
\endhead

\midrule
\multicolumn{3}{r}{\textit{Continued on next page}} \\
\endfoot

\bottomrule
\endlastfoot

GCN & Reduce function & $\{\texttt{sum},\ \texttt{max},\ \texttt{mean}\}$ \\
 & Message function & \texttt{SAFE\_MSG\_FUNCS} (DGL message functions) \\
 & Activation & $\{\texttt{relu},\ \texttt{elu},\ \texttt{gelu},\ \texttt{leaky\_relu},\ \texttt{silu}\}$ \\
 & Bias & $\{\texttt{True},\ \texttt{False}\}$ \\
 & \#Layers & $\{1,\dots,4\}$ \\
 & Hidden sizes & layer2: 8--127; layer3: 8--63; layer4: 8--31 \\
 & Balancing strategy & $\{\texttt{class\_weights},\ \texttt{oversampling},\ \texttt{none}\}$ \\
 & Oversample ratio & uniform $[0.1,\ 1.0]$ \\
 & Dropout rate & uniform $[0.0,\ 0.3]$ \\
 & Learning rate & log-uniform $[10^{-4},\ 5\times10^{-2}]$ \\
 & Monotonic width constraints &
for $\geq$3 layers: layer3 $<$ layer2;\newline
for $\geq$4 layers: layer4 $<$ layer3 \\
\midrule

GAT & Optimizer &
$\{\texttt{Adam},\allowbreak \texttt{SGD},\allowbreak \texttt{Adadelta},\allowbreak
\texttt{Adagrad},\allowbreak \texttt{AdamW},\allowbreak \texttt{Adamax},\allowbreak
\texttt{ASGD},\allowbreak \texttt{RMSprop},\allowbreak \texttt{Rprop}\}$ \\
 & Activation & $\{\texttt{relu},\ \texttt{elu},\ \texttt{gelu},\ \texttt{leaky\_relu},\ \texttt{silu}\}$ \\
 & \#Layers & $\{1,\dots,4\}$ \\
 & Hidden sizes & layer2: 8--127; layer3: 8--63; layer4: 8--31 \\
 & Balancing strategy & $\{\texttt{class\_weights},\ \texttt{none}\}$ \\
 & Dropout rate (post-layer) & uniform $[0.0,\ 0.3]$ \\
 & Learning rate & log-uniform $[10^{-4},\ 5\times10^{-2}]$ \\
 & Attention heads & $\{1,\ 2,\ 4,\ 8\}$ \\
 & LeakyReLU negative slope & uniform $[0.01,\ 0.3]$ \\
 & Residual & $\{\texttt{True},\ \texttt{False}\}$ \\
 & Attention dropout & uniform $[0.0,\ 0.4]$ \\
 & Feature dropout & uniform $[0.0,\ 0.4]$ \\
 & Monotonic width constraints &
for $\geq$3 layers: layer3 $<$ layer2;\newline
for $\geq$4 layers: layer4 $<$ layer3 \\
\midrule

GCL & hidden\_dim & $\{32,\ 64,\ 128\}$ \\
 & learning\_rate\_pretrain & log-uniform $[5\times10^{-4},\ 5\times10^{-3}]$ \\
 & edge\_drop\_prob & uniform $[0.1,\ 0.4]$ \\
 & learning\_rate\_classifier & log-uniform $[5\times10^{-4},\ 5\times10^{-2}]$ \\
 & weight\_decay\_classifier & log-uniform $[10^{-6},\ 10^{-2}]$ \\
 & class\_weight\_base & log-uniform $[0.5,\ 5.0]$ \\
 & class\_weight\_exp & log-uniform $[0.5,\ 3.0]$ \\
 & encoder\_lr\_factor & $\{0.05,\ 0.1,\ 0.2,\ 0.3\}$ \\
\end{longtable}

All hyperparameters are optimized via Bayesian optimization (Hyperopt) with a fixed trial budget.

\section{Peak Performance Gains per Graph Signal Category}
\label{app:PeakPerformanceGainsperGraphSignalCategory}

This figure complements the average performance analysis in Section 8.2 by illustrating the best-case performance attainable per graph signal category across all classifiers. 

\begin{figure}[H]
\centering
\includegraphics[width=\linewidth]{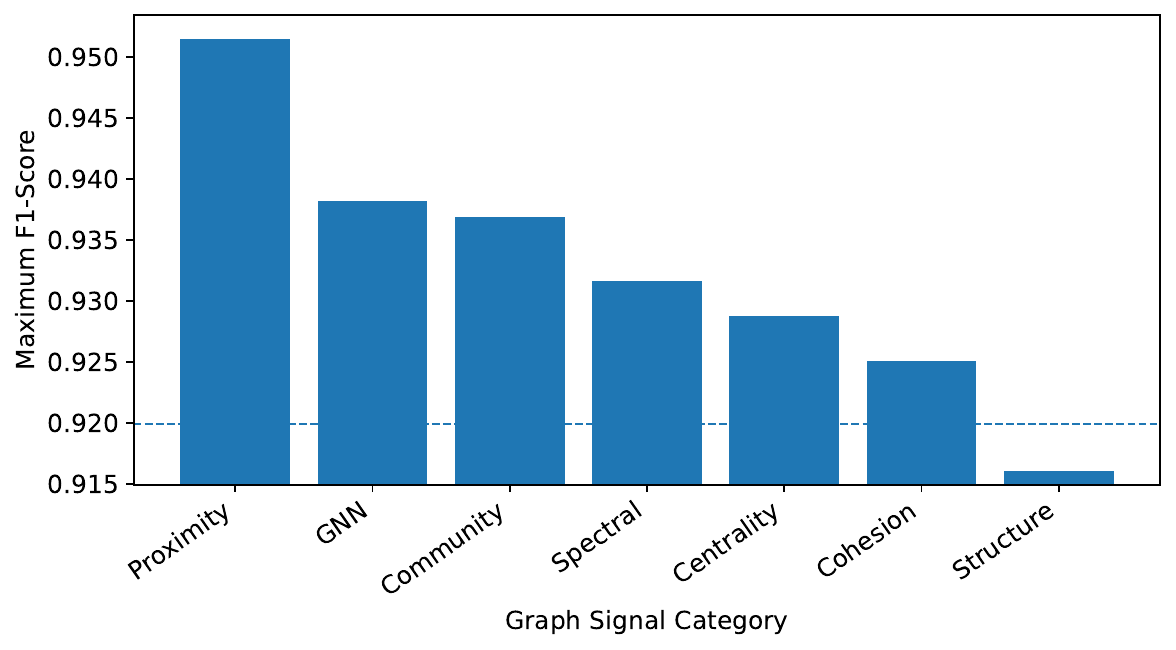} 
\caption{Maximum $F_1$-score achieved per graph signal category across all evaluated classifiers. The dashed horizontal line denotes the best-performing transaction-only (TRX) baseline.}
\label{fig:bar_category_max_f1}
\end{figure}




\bibliographystyle{plain}
\bibliography{references}


%


\end{document}